\documentclass[11pt]{article}

\usepackage[preprint]{acl}

\usepackage{times}
\usepackage{amsfonts}
\usepackage{latexsym}
\usepackage[T1]{fontenc}

\usepackage{stfloats}
\usepackage[utf8]{inputenc}
\usepackage{array} 
\usepackage[most]{tcolorbox}
\usepackage{cleveref}
\usepackage[table]{xcolor}
\usepackage{amsmath}
\usepackage{microtype}
\usepackage{booktabs,tabularx,ragged2e,array}
\usepackage{inconsolata}
\usepackage{siunitx}
\usepackage{tikz}

\usepackage{graphicx}
\usepackage{pifont}
\usepackage{makecell}
\usepackage{multirow}
\usepackage{kotex}
\usepackage{adjustbox}
\usepackage{listings}

\usepackage{colortbl}        
\usepackage{pgf}             
\usepackage[normalem]{ulem}  
\usepackage{comment}
\usepackage{enumitem}
\lstdefinestyle{promptstyle}{
  basicstyle=\ttfamily\scriptsize,
  columns=fullflexible,
  breaklines=true,
  keepspaces=true,
  showstringspaces=false,
  aboveskip=0pt,
  belowskip=0pt
}

\newtcblisting{promptlisting}{
  listing engine=listings,
  listing only,
  listing options={style=promptstyle},
  enhanced,
  colback=black!2,
  colframe=black!45,
  boxrule=0.6pt,
  arc=2pt,
  left=7pt,right=7pt,top=7pt,bottom=7pt,
}

\DeclareUnicodeCharacter{266B}{}

\newcommand{\best}[1]{\textbf{#1}}
\newcommand{\second}[1]{\underline{#1}}
\definecolor{llmshade}{gray}{0.92}  
\definecolor{llmhead}{gray}{0.87}   

\definecolor{overlap}{RGB}{255,245,170}

\newcolumntype{Y}{>{\RaggedRight\arraybackslash}X}

\definecolor{overlapred}{RGB}{255,215,215} 
\newcommand{\rhl}[1]{\begingroup\setlength{\fboxsep}{0.6pt}\colorbox{overlapred}{\strut #1}\endgroup}

\title{Enhancing Multilingual RAG Systems with Debiased Language Preference-Guided Query Fusion}

\author{
Jeonghyun Park,~Byeongjeong Kim,~Seojin Hwang,~Hwanhee Lee\thanks{Corresponding author.} \\
    Chung-Ang University, Seoul, Korea\\
    \texttt{\{tom0365, michael97k, swiftie1230, hwanheelee\}@cau.ac.kr} \\
\url{https://jeonghyunpark2002.github.io/DELTA_project_page}
}

\begin{document}
\maketitle

\begin{abstract}
Multilingual Retrieval-Augmented Generation (mRAG) systems often exhibit a perceived preference for high-resource languages, particularly English, resulting in the widespread adoption of English pivoting. While prior studies attribute this advantage to the superior English-centric capabilities of Large Language Models (LLMs), we find that such measurements are significantly distorted by structural priors inherent in evaluation benchmarks. 
Specifically, we identify \textit{exposure bias} and a \textit{gold availability prior}—both driven by the disproportionate concentration of resources in English—as well as \textit{cultural priors} rooted in topic locality, as factors that hinder accurate assessment of genuine language preference.
To address these biases, we propose \textbf{DeLP} (\textbf{De}biased \textbf{L}anguage \textbf{P}reference), a calibrated metric designed to explicitly factor out these structural confounds. Our analysis using DeLP reveals that the previously reported English preference is largely a byproduct of evidence distribution rather than an inherent model bias. Instead, we find that retrievers fundamentally favor monolingual alignment between the query and the document language.
Building on this insight, we introduce \textbf{DELTA} (\textbf{DE}biased \textbf{L}anguage preference–guided \textbf{T}ext \textbf{A}ugmentation), a lightweight and efficient mRAG framework that strategically leverages monolingual alignment to optimize cross-lingual retrieval and generation. 
Experimental results demonstrate that DELTA consistently outperforms English pivoting and mRAG baselines across diverse languages. The Code is available at \url{https://github.com/jeonghyunpark2002/DELTA.git}

\end{abstract}

\section{Introduction}
Multilingual Retrieval-Augmented Generation (mRAG)~\cite{chirkova-etal-2024-retrieval} generalizes Retrieval-Augmented Generation (RAG)~\cite{lewis2020retrieval} by retrieving evidence from multilingual knowledge sources. This enables language models to produce responses that are not only factually grounded but also sensitive to the user’s language and linguistic context~\cite{rau2024bergen}.
In this landscape, mRAG systems frequently exhibit a significant language preference for English~\cite{zhang2023don, park2025investigating}. Consequently, English pivoting—the practice of translating a non-English query into English before retrieval—has emerged as a surprisingly strong heuristic that yields substantial gains across many languages~\cite{chirkova-etal-2024-retrieval, ranaldi2025multilingual}. Prior works have largely attributed this advantage to the "English-centric" competence of generators, such as superior reasoning in English or reduced translation noise, leading to research that primarily intervenes at the generation stage~\cite{chirkova-etal-2024-retrieval, li2025language, moon2025quality}.


\begin{figure}[t]

  \centering
  \includegraphics[width=1.0\columnwidth]{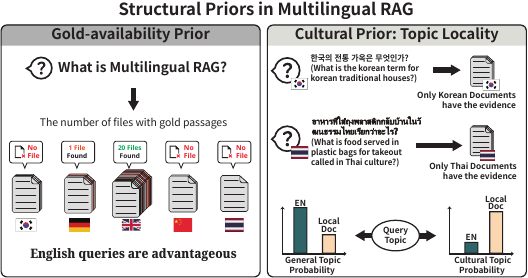}
  \caption{Common causes of language preference of mRAG: gold-availability prior and cultural prior.}
  \label{fig:motivation}
  \vspace{-3mm}
\end{figure}

However, we find that the perceived effectiveness of English pivoting is primarily driven by retrieval-side structural biases rather than any inherent linguistic preference of the model. As illustrated in Figure~\ref{fig:motivation}, we identify two major confounders: a \textbf{gold availability prior} and \textbf{cultural priors}. Our analysis shows that ground-truth evidence in standard benchmarks is overwhelmingly concentrated in English resources, establishing a dominant gold availability prior. This concentration not only makes English the sole or primary location where correct evidence exists, but also leads to \textbf{exposure bias}---the retrieval 
system's inherent tendency to surface documents from more prevalent languages regardless of the query's intended language---of English documents during retrieval, further amplifying English's advantage.
Together, these effects fundamentally distort measured language preference and inflate the apparent superiority of English. In addition, we identify \textbf{cultural priors} as an equally critical factor. There are benchmark questions that are tied to specific geographic or cultural contexts and contain native surface forms (e.g., local titles, aliases, and scripts) that act as strong retrieval anchors. When benchmarks overrepresent such locale-specific topics, languages may appear “preferred” due to query–corpus alignment and environmental exposure rather than the model’s intrinsic preference. Critically, these structural factors contaminate existing methods~\cite{park2025investigating} for measuring language preference.

To reveal the intrinsic preference of mRAG systems, we propose Debiased Language Preference (\textbf{DeLP}), a calibrated measurement that explicitly regresses out these structural confounds. DeLP utilizes a ridge regression framework to predict observed language preference from structural factors (e.g., corpus size, gold availability, cultural prior), treating the residual signal as the true, debiased preference of the model. By applying DeLP, we reveal a qualitatively different landscape: the previously inflated preference for English largely evaporates. Instead, our results show an increased preference for \textbf{monolingual alignment}, where the retriever performs most effectively when the query and the target document languages match.


Building on the discovery of monolingual alignment, we introduce DEbiased Language preference–guided Text Augmentation (\textbf{DELTA}), a lightweight query-level solution for mRAG. 
DELTA leverages the debiased preference signals from DeLP to dynamically identify intrinsic model preference for a given query, effectively bridging the gap between the user's query and the languages where the model performs most reliably.
By reformulating the query to include these preference-aligned multilingual anchors, DELTA preserves the native script's context while maximizing the benefits of monolingual alignment. DELTA is highly cost-effective, requiring no modifications to the underlying corpus or retriever architecture. Our experiments demonstrate that DELTA outperforms naive English pivoting, proving that accounting for the model's true linguistic preference—rather than following biased environmental cues—is the key to unlocking the true potential of mRAG systems.
\section{The Myth of English Preference: Structural Priors in mRAG}
\label{sec:prior}
A dominant mRAG strategy is \textit{English pivoting}, where non-English queries are translated into English to exploit the perceived superiority of English-centric models. We hypothesize that these gains are not necessarily indicative of model preference but rather reflect a massive exposure bias rooted in the structural distribution of evidence.

\newcommand{\enmark}{\textcolor{red}{\scriptsize(\,EN\,)}}

\begin{table*}[t]
\centering
\small
\setlength{\tabcolsep}{2.5pt}
\renewcommand{\arraystretch}{1.10}

\begin{adjustbox}{max width=\textwidth} 
\begin{tabular}{l|rr|
r@{\hspace{0.6pt}}r|r@{\hspace{0.6pt}}r|r@{\hspace{0.6pt}}r|r@{\hspace{0.6pt}}r}
\toprule
& \multicolumn{2}{c|}{Gold Availability} 
& \multicolumn{2}{c|}{Retriever Recall}
& \multicolumn{2}{c|}{Qwen3-235B-A22B}
& \multicolumn{2}{c|}{Gemini-2.5-Flash}
& \multicolumn{2}{c}{DeepSeek-Chat-v3.1} \\
\midrule
\textbf{lang}
& \textbf{\#q} & \textbf{ratio}
& \textbf{Base} & \textbf{EN}
& \textbf{Base} & \textbf{EN}
& \textbf{Base} & \textbf{EN}
& \textbf{Base} & \textbf{EN} \\
\midrule
en
& 26934 & \cellcolor{red!55}\textbf{73.29\%}
& -- & --
& 70.05\enmark & --
& 58.26\enmark & --
& 60.77\enmark & -- \\
ar
& 214 & \cellcolor{red!8}0.58\%
& 13.36 & 23.57
& 47.79 & 55.14
& 40.79 & 48.44
& 43.64 & 50.97 \\
de
& 435 & \cellcolor{red!10}1.18\%
& 21.62 & 26.40
& 63.81 & 60.72
& 53.52 & 55.17
& 54.16 & 56.92 \\
ja
& 513 & \cellcolor{red!11}1.40\%
& 16.84 & 25.83
& 46.60 & 59.29
& 44.26 & 53.68
& 44.72 & 56.52 \\
ko
& 306 & \cellcolor{red!9}0.83\%
& 15.62 & 24.81
& 40.14 & 54.57
& 35.97 & 47.67
& 34.21 & 50.49 \\
th
& 187 & \cellcolor{red!8}0.51\%
& 21.90 & 27.55
& 40.73 & 60.46
& 31.65 & 54.86
& 36.80 & 57.52 \\
zh
& 287 & \cellcolor{red!9}0.78\%
& 16.47 & 26.53
& 37.52 & 59.53
& 30.81 & 53.59
& 33.14 & 56.11 \\
\bottomrule
\end{tabular}
\end{adjustbox}
\vspace{-2mm}
\caption{Gold availability bias and its impact on multilingual RAG.
Gold Availability measures gold-passage coverage per language, and Retriever Recall reports Recall@50. Model columns show end-to-end accuracy. Base denotes native-language queries, and EN denotes English-translated queries.}

\label{tab:rq1_single_table_nodelta}
\vspace{-6mm}
\end{table*}

\subsection{Experimental Setup}
\paragraph{Datasets}
We conduct our analysis on MKQA~\cite{longpre-etal-2021-mkqa}, which provides 10k professionally translated queries. To enable precise measurement of evidence location, we use a 2.7K-example subset that overlaps with KILT NQ~\footnote{\url{https://huggingface.co/datasets/facebook/kilt_tasks}}. Since MKQA does not provide standardized provenance for each translated instance, using KILT allows us to inherit document-level provenance (i.e., gold Wikipedia passage IDs), which is essential for quantifying gold availability across different linguistic corpora.

\vspace{-2mm}
\paragraph{Models and Knowledge Sources}
We employ BGE-m3~\cite{bge-m3} as the multilingual retriever and re-ranker. For the generation, we use three recently released robust multilingual LLMs: Qwen3-235B~\cite{yang2025qwen3}, DeepSeek-v3.1~\cite{liu2024deepseek}, and Gemini-2.5-Flash~\cite{comanici2025gemini}. We retrieve top-50 candidate documents per query and apply re-ranking, using the top-5 documents as contexts for generation. In line with previous work~\cite{chirkova-etal-2024-retrieval, park2025investigating}, we use Wikipedia editions in English and the user's local language to serve as the knowledge sources. Detailed corpus statistics are provided in Appendix~\ref{app:statistics}.

\subsection{Linguistic Superiority or Data Imbalance?}
\label{sec:gold}


Following the MKQA protocol anchored to KILT~\cite{longpre-etal-2021-mkqa}, we identify the location of gold passages (WPIDs) within the multilingual Wikipedia datastore.
We report this distribution as Gold Availability, as the number of queries whose gold passage WPID is present in that language’s Wikipedia corpus for each language of query.
This distribution reflects the extent of corpus-level coverage of gold evidence in each language within the benchmark. We then relate this to retrieval performance, measured by Recall@50—i.e., the fraction of queries whose gold passages appear within the top-50 retrieved candidates—under both native-language queries and English queries (EN). We further evaluate end-to-end mRAG performance using character 3-gram recall between the generated and reference answers.
Details are provided in Appendix~\ref{app:gold}.



Our analysis in Table~\ref{tab:rq1_single_table_nodelta} reveals an extreme imbalance in the retrieval environment. English Wikipedia provides substantially higher document density and coverage, introducing a strong \textbf{exposure bias}. More critically, for a vast majority of queries, English Wikipedia also serves as the sole repository of ground-truth, 
inducing a dominant \textbf{gold availability prior}. 
Consequently, English pivoting appears effective not because models prefer English, but because of this structural skew, 
sustaining the long-standing "myth" of English preference.



\begin{table}[!htbp]
\centering
\scriptsize
\setlength{\tabcolsep}{2.5pt}
\renewcommand{\arraystretch}{1.15}

\begin{tabularx}{\columnwidth}{@{}*{8}{>{\centering\arraybackslash}X}@{}}
\toprule
\textbf{ar} & \textbf{de} & \textbf{es} & \textbf{fr} & \textbf{ja} & \textbf{ko} & \textbf{ru} & \textbf{zh} \\
\midrule
21.43\% & 16.67\% & 18.52\% & 21.74\% & 14.29\% & 12.50\% & 25.00\% & 6.67\% \\
\bottomrule
\end{tabularx}

\vspace{-2mm}
\caption{Local-gold coverage by predicted $L_{\mathrm{loc}}$.}
\label{tab:cultural_prior_mo}
\vspace{-3mm}
\end{table}

\vspace{-3mm}
\subsection{Impact of Cultural Priors}
\label{sec:cultural_mo}
Queries often carry cultural or regional context, whose associated language can naturally align with local-language evidence and be conflated with language preference; we therefore examine where their gold documents reside across Wikipedia languages.
We first isolate queries that involve cultural or regional contexts by instructing GPT-4o-mini~\cite{hurst2024gpt} to predict a single primary locale language $L_{\mathrm{loc}}$, selecting the language that corresponds to the query’s main referenced region or culture (Details on the classifier are in Appendix~\ref{app:cultural}).
Table~\ref{tab:cultural_prior_mo} reports the local-gold rate $p(\text{gold WPID exists in local Wikipedia}\mid L_{\mathrm{loc}})$ for each predicted language. 
Local evidence is not uniformly absent—across several predicted locale languages, about 20\% of these queries have gold pages only in the corresponding local Wikipedia. 
This distribution introduces a structural bias for retrieval to rely on locale-specific surface-form anchors (e.g., native titles, aliases, scripts) when they exist. As a result, observed language preference can be influenced by locale-tied queries and their local-gold presence, motivating an explicit cultural prior term $p_{\mathrm{cult}}$ to avoid conflating topic locality. 

\begin{figure*}[t]
  \centering
  \includegraphics[width=0.98\textwidth]{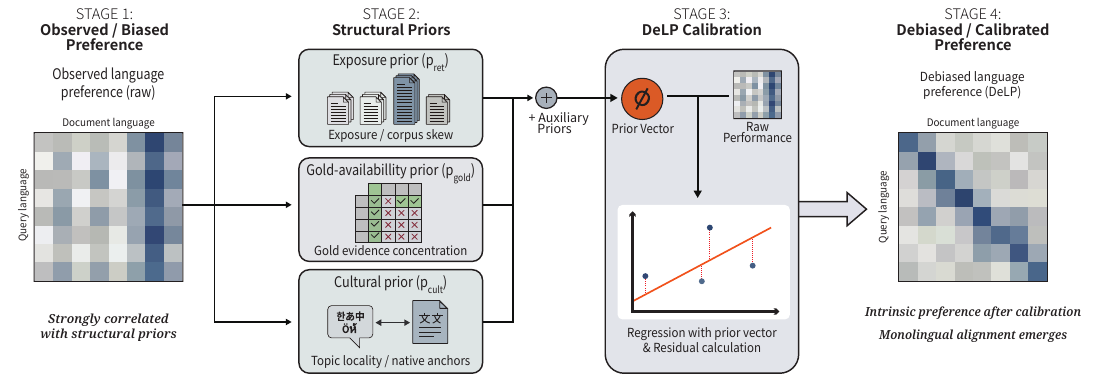}
  \vspace{-3mm}
  \caption{Overview of DeLP, which measures intrinsic language preference in mRAG by regressing out exposure, gold-availability, and cultural priors from raw preference signals.}
  \label{fig:delp}
  \vspace{-6mm}
\end{figure*}
\section{Measuring Language Preference via Bias Calibration}
\label{sec:calibration}

The structural priors identified in Section~\ref{sec:prior} suggest that existing metrics, such as MLRS~\cite{park2025investigating}, fail to distinguish between a model's intent to use a language and the external necessity imposed by data distribution. To reveal the genuine preference, we introduce \textbf{De}biased \textbf{L}anguage \textbf{P}reference (\textbf{DeLP}), a calibrated measurement framework that explicitly regresses out structural confounds.

\subsection{Decomposing Structural Bias in mRAG}
To isolate intrinsic model preference, we decompose the confounding factors identified in our previous analysis into three primary priors. 

\vspace{-2mm}
\paragraph{Exposure prior ($p_{\mathrm{ret}}$).}
As observed in the exposure bias of Section~\ref{sec:gold}, high-resource corpora—particularly English—dominate the top retrieval results regardless of the encoder's linguistic intent. This prior captures the "popularity bias" of the datastore. A language that appears more frequently in the candidate pool is more likely to be retrieved, potentially leading to a false inflation of preference. Let $L_q$ denote the query language and $L_d$ the language of a retrieved document. We estimate $p_{\mathrm{ret}}(L_d \mid L_q)$ by calculating the average proportion of document language $L_d$ within the top-50 candidates for queries in $L_q$.
\vspace{-2mm}
\paragraph{Gold-availability prior ($p_{\mathrm{gold}}$).}
Our findings in Table~\ref{tab:rq1_single_table_nodelta} (Section~\ref{sec:gold}) demonstrate that retrieval is often forced into English because the gold evidence simply does not exist elsewhere. To prevent such uncontrollable circumstances from being mistaken for model preference, we explicitly model the availability of ground-truth passages. We estimate $p_{\mathrm{gold}}(L_q, L_d)$ on the MKQA--KILT overlap as the empirical fraction of queries in $L_q$ for which gold evidence is present in the $L_d$ corpus.

\paragraph{Cultural prior ($p_{\mathrm{cult}}$).}
As discussed in Section~\ref{sec:cultural_mo}, locale-tied queries contain native surface forms that act as structural anchors, naturally pulling retrieval toward the corresponding local-language evidence. This prior captures topic locality; a retriever might show high preference scores for a language simply because the topic is regional. We estimate $p_{\mathrm{cult}}(L_d)$ by identifying the query's associated locale language $L_{\mathrm{loc}}$ using \textit{GPT-4o-mini} and computing the fraction of queries where $L_{\mathrm{loc}} = L_d$. The classifier selects the local language of the primary referenced place or culture (e.g., "When did Hong Kong go back to China?" → \textit{zh}), reserving en for inherently English-speaking or genuinely global queries. For the details of how we measure cultural prior, please refer to Appendix~\ref{app:cultural}.

\subsection{Calibration for Genuine Preference}

We calibrate the raw language preference scores with respect to the above priors to obtain the residual signal-the component not explained by the priors, which reflects the model’s genuine language preference.
In addition to the above three main priors, we incorporate two auxiliary structural controls, namely a corpus-size prior $p_{\mathrm{db}}$ and a passage-length statistic $\ell$, as additional covariates in the prior feature vector $\phi(L_q,L_d)$ (defined below) to account for language-dependent corpus scale and length effects.
Let $s_e(L_q, L_d)$ denote the observed language-preference score of the encoder $e$ for the query language $L_q$, and for the language of evidence $L_d$.
We instantiate $s_e$ by MLRS~\cite{park2025investigating} (Table~\ref{tab:mlrs}), which measures how often the retriever surfaces evidence in each $L_d$ for a fixed $L_q$. For each language pair $(L_q, L_d)$, we define a prior feature vector $\phi(L_q, L_d) \in \mathbb{R}^7$:

\definecolor{heat}{RGB}{255,220,80}    

\newcommand{\deltasize}{\fontsize{5.4}{5.8}\selectfont}

\newcommand{\rowmin}{0}
\newcommand{\rowmax}{1}
\newcommand{\setrowrange}[2]{\gdef\rowmin{#1}\gdef\rowmax{#2}}

\newcommand{\heatcell}[2]{%
  \begingroup
  \pgfmathsetmacro{\val}{#1}%
  \pgfmathsetmacro{\minval}{\rowmin}%
  \pgfmathsetmacro{\maxval}{\rowmax}%
  \pgfmathsetmacro{\maxint}{65}
  \pgfmathsetmacro{\den}{(\maxval-\minval)}%
  \pgfmathparse{ifthenelse(abs(\den)<0.000001,0,\maxint*(\val-\minval)/\den)}%
  \pgfmathsetmacro{\raw}{\pgfmathresult}%
  \pgfmathsetmacro{\clamped}{max(0,min(\maxint,\raw))}%
  \pgfmathtruncatemacro{\heatpct}{\clamped}%
  \xdef\heatpct{\heatpct}%
  \cellcolor{heat!\heatpct}#2%
  \endgroup
}

\begin{table*}[t]
\centering
\fontsize{7.2}{8.4}\selectfont
\setlength{\tabcolsep}{2.0pt}
\renewcommand{\arraystretch}{1.03}

{\setlength{\arrayrulewidth}{0.3pt}%
\begin{tabular}{@{}c c|*{8}{c}@{}}
\toprule
\multirow{2}{*}{\textbf{\shortstack{Query\\Lang.}}} &
\multirow{2}{*}{$\mathbf{L_q=L_d}$} &
\multicolumn{8}{|c}{$\mathbf{L_q \neq L_d}$} \\
\cmidrule(lr){3-10}
 &  & \textbf{\textit{en}} & \textbf{\textit{ko}} & \textbf{\textit{zh}} & \textbf{\textit{fr}} & \textbf{\textit{ja}} & \textbf{\textit{it}} & \textbf{\textit{pt}} & \textbf{\textit{es}} \\
\midrule

\setrowrange{35.49}{50.10}
\textbf{\textit{en}} &
\heatcell{50.10}{\best{50.10} \textcolor{gray}{\deltasize(56.79)}} &
\textcolor{gray}{--} &
\heatcell{36.67}{36.67 \textcolor{gray}{\deltasize(33.94)}} &
\heatcell{40.34}{\second{40.34} \textcolor{gray}{\deltasize(33.99)}} &
\heatcell{35.57}{35.57 \textcolor{gray}{\deltasize(37.57)}} &
\heatcell{39.61}{39.61 \textcolor{gray}{\deltasize(34.18)}} &
\heatcell{35.49}{35.49 \textcolor{gray}{\deltasize(36.79)}} &
\heatcell{36.46}{36.46 \textcolor{gray}{\deltasize(36.54)}} &
\heatcell{36.02}{36.02 \textcolor{gray}{\deltasize(37.49)}} \\
\midrule

\setrowrange{34.70}{43.59}
\textbf{\textit{ko}} &
\heatcell{43.38}{\second{43.38} \textcolor{gray}{\deltasize(42.21)}} &
\heatcell{37.69}{37.69 \textcolor{gray}{\deltasize(44.36)}} &
\textcolor{gray}{--} &
\heatcell{41.75}{41.75 \textcolor{gray}{\deltasize(35.44)}} &
\heatcell{34.90}{34.90 \textcolor{gray}{\deltasize(36.84)}} &
\heatcell{43.59}{\best{43.59} \textcolor{gray}{\deltasize(38.22)}} &
\heatcell{34.70}{34.70 \textcolor{gray}{\deltasize(36.00)}} &
\heatcell{35.65}{35.65 \textcolor{gray}{\deltasize(35.71)}} &
\heatcell{34.77}{34.77 \textcolor{gray}{\deltasize(36.24)}} \\
\midrule

\setrowrange{34.73}{50.60}
\textbf{\textit{zh}} &
\heatcell{50.60}{\best{50.60} \textcolor{gray}{\deltasize(45.81)}} &
\heatcell{38.68}{38.68 \textcolor{gray}{\deltasize(45.35)}} &
\heatcell{37.95}{37.95 \textcolor{gray}{\deltasize(35.06)}} &
\textcolor{gray}{--} &
\heatcell{34.73}{34.73 \textcolor{gray}{\deltasize(36.73)}} &
\heatcell{41.90}{\second{41.90} \textcolor{gray}{\deltasize(36.51)}} &
\heatcell{34.87}{34.87 \textcolor{gray}{\deltasize(36.21)}} &
\heatcell{35.84}{35.84 \textcolor{gray}{\deltasize(35.91)}} &
\heatcell{35.22}{35.22 \textcolor{gray}{\deltasize(36.69)}} \\
\midrule

\setrowrange{36.00}{41.16}
\textbf{\textit{fr}} &
\heatcell{40.05}{40.05 \textcolor{gray}{\deltasize(43.74)}} &
\heatcell{41.16}{\best{41.16} \textcolor{gray}{\deltasize(47.84)}} &
\heatcell{36.77}{36.77 \textcolor{gray}{\deltasize(34.03)}} &
\heatcell{40.50}{\second{40.50} \textcolor{gray}{\deltasize(34.16)}} &
\textcolor{gray}{--} &
\heatcell{39.92}{39.92 \textcolor{gray}{\deltasize(34.50)}} &
\heatcell{36.00}{36.00 \textcolor{gray}{\deltasize(37.31)}} &
\heatcell{36.69}{36.69 \textcolor{gray}{\deltasize(36.76)}} &
\heatcell{36.28}{36.28 \textcolor{gray}{\deltasize(37.76)}} \\
\midrule

\setrowrange{34.94}{49.19}
\textbf{\textit{ja}} &
\heatcell{49.19}{\best{49.19} \textcolor{gray}{\deltasize(45.50)}} &
\heatcell{38.70}{38.70 \textcolor{gray}{\deltasize(45.37)}} &
\heatcell{38.40}{38.40 \textcolor{gray}{\deltasize(35.69)}} &
\heatcell{41.56}{\second{41.56} \textcolor{gray}{\deltasize(35.24)}} &
\heatcell{34.94}{34.94 \textcolor{gray}{\deltasize(36.94)}} &
\textcolor{gray}{--} &
\heatcell{34.99}{34.99 \textcolor{gray}{\deltasize(36.29)}} &
\heatcell{35.97}{35.97 \textcolor{gray}{\deltasize(36.04)}} &
\heatcell{35.23}{35.23 \textcolor{gray}{\deltasize(36.70)}} \\
\midrule

\setrowrange{36.63}{40.63}
\textbf{\textit{it}} &
\heatcell{39.05}{39.05 \textcolor{gray}{\deltasize(41.72)}} &
\heatcell{40.63}{\best{40.63} \textcolor{gray}{\deltasize(47.30)}} &
\heatcell{36.85}{36.85 \textcolor{gray}{\deltasize(34.12)}} &
\heatcell{40.59}{\second{40.59} \textcolor{gray}{\deltasize(34.25)}} &
\heatcell{36.63}{36.63 \textcolor{gray}{\deltasize(38.64)}} &
\heatcell{39.86}{39.86 \textcolor{gray}{\deltasize(34.44)}} &
\textcolor{gray}{--} &
\heatcell{37.01}{37.01 \textcolor{gray}{\deltasize(37.09)}} &
\heatcell{36.91}{36.91 \textcolor{gray}{\deltasize(38.39)}} \\
\midrule

\setrowrange{36.48}{46.08}
\textbf{\textit{pt}} &
\heatcell{46.08}{\best{46.08} \textcolor{gray}{\deltasize(39.76)}} &
\heatcell{40.55}{40.55 \textcolor{gray}{\deltasize(47.23)}} &
\heatcell{36.98}{36.98 \textcolor{gray}{\deltasize(34.24)}} &
\heatcell{40.63}{\second{40.63} \textcolor{gray}{\deltasize(34.29)}} &
\heatcell{36.50}{36.50 \textcolor{gray}{\deltasize(38.52)}} &
\heatcell{40.01}{40.01 \textcolor{gray}{\deltasize(34.59)}} &
\heatcell{36.48}{36.48 \textcolor{gray}{\deltasize(37.80)}} &
\textcolor{gray}{--} &
\heatcell{37.73}{37.73 \textcolor{gray}{\deltasize(39.21)}} \\
\midrule

\setrowrange{36.30}{40.71}
\textbf{\textit{es}} &
\heatcell{38.19}{38.19 \textcolor{gray}{\deltasize(41.30)}} &
\heatcell{40.71}{\best{40.71} \textcolor{gray}{\deltasize(47.39)}} &
\heatcell{36.86}{36.86 \textcolor{gray}{\deltasize(34.13)}} &
\heatcell{40.39}{\second{40.39} \textcolor{gray}{\deltasize(34.04)}} &
\heatcell{36.30}{36.30 \textcolor{gray}{\deltasize(38.31)}} &
\heatcell{39.76}{39.76 \textcolor{gray}{\deltasize(34.34)}} &
\heatcell{36.45}{36.45 \textcolor{gray}{\deltasize(37.76)}} &
\heatcell{37.25}{37.25 \textcolor{gray}{\deltasize(37.32)}} &
\textcolor{gray}{--} \\
\bottomrule
\end{tabular}
} 

\vspace{-1.5mm}
\caption{DeLP for query-document language pairs, averaged over three encoders (raw MLRS in parentheses). Background shading is row-wise min-max scaled (darker = stronger preference); dashes denote $L_q = L_d$. Underline denotes the second-highest per row. The relatively stronger English and Chinese signals are attributable to encoder training-data language imbalance (e.g., BGE-m3 trains on 194 languages with English 43.9\% and Chinese 20.5\%).}
\vspace{-5mm}
\label{tab:delp}
\end{table*}

\begingroup
\small
\renewcommand{\arraystretch}{0.92}
\begin{equation}
\label{eq:phi}
\phi(L_q,L_d)=
\begin{bmatrix}
1 \\
\log\!\big(p_{\mathrm{ret}}(L_d\!\mid\!L_q)+\epsilon\big) \\
\log\!\big(p_{\mathrm{db}}(L_d)+\epsilon\big) \\
\log\!\big(\ell(L_d)+\epsilon\big) \\
\log\!\big(p_{\mathrm{gold}}(L_q,L_d)+\epsilon\big) \\
\log\!\big(p_{\mathrm{cult}}(L_d)+\epsilon\big) \\
\mathbb{I}[L_q = L_d]
\end{bmatrix}.
\end{equation}
\endgroup

where $p_{\mathrm{ret}}$ is the exposure prior, $p_{\mathrm{db}}$ is the corpus-size prior,
$\ell$ is a passage-length statistic (e.g., median length), $p_{\mathrm{gold}}$ is the gold-availability prior, and $p_{\mathrm{cult}}$ is the cultural prior, and $\epsilon > 0$ is a small constant added for numerical stability in the log transformation.
The vector $\phi(L_q,L_d)$ stacks interpretable covariates that predict $s_e$ without invoking intrinsic model preference. We use log-transformed priors to compress heavy-tailed probabilities and corpus statistics, and make linear effects more reasonable across languages.
The indicator $\mathbb{I}[L_q = L_d]$ allows the model to treat same-language retrieval as a special case, ensuring that monolingual matching is not forced to be explained solely by external priors.

\paragraph{Ridge calibration.}
We fit the regression separately for each encoder $e$ to learn how much of its observed score $s_e$ can be attributed to structural priors.
We use ridge regularization to stabilize coefficients under the various priors, preventing any single feature from disproportionately absorbing the preference signal.
Let $\mathcal{C}$ be the set of all language pairs used for calibration. For each encoder $e$, we fit a ridge regression that predicts the raw score from priors:

\begingroup
\small
\setlength{\abovedisplayskip}{4pt}
\setlength{\belowdisplayskip}{4pt}
\begin{equation}
\label{eq:ridge}
\begin{aligned}
\hat{\beta}_e &= \arg\min_{\beta}\; \mathcal{J}_e(\beta),\\
\mathcal{J}_e(\beta) &= \sum_{(L_q,L_d)\in\mathcal{C}}
\big(s_e(L_q,L_d)-\phi(L_q,L_d)^{\top}\beta\big)^2 \\
&\quad + \lambda \lVert \beta \rVert_2^2 .
\end{aligned}
\end{equation}

\endgroup
where $\lambda$ is a regularization hyperparameter and $\epsilon$ is a small constant for numerical stability. 
 
\paragraph{Debiased preference (DeLP).}
We define the debiased preference as the residual signal after removing the component explained by structural priors:
\par\begingroup\small
\begin{equation}
\label{eq:resid}
r_e(L_q,L_d)=s_e(L_q,L_d)-\phi(L_q,L_d)^{\!\top}\hat{\beta}_e .
\end{equation}
\endgroup\par
The residual $r_e(L_q,L_d)$ represents the portion of the observed score that is independent of structural priors.
To keep the overall scale comparable to the raw score, we re-center the residuals by the global mean of raw scores $\mu_e$:
\begin{equation}
\label{eq:delp}
\begin{aligned}
\mathrm{DeLP}_e(L_q,L_d)
&= r_e(L_q,L_d) + \mu_e, \\
\mu_e
&= \frac{1}{|\mathcal{C}|}
\sum_{(L_q,L_d)\in\mathcal{C}} s_e(L_q,L_d).
\end{aligned}
\end{equation}
By adding back $\mu_e$, DeLP stays on a numeric scale comparable to standard MLRS tables while preserving the relative differences that define the model's intrinsic tendencies. 
To mitigate potential encoder-specific bias, we apply our calibration procedure independently to each retriever and report all debiased results for three multilingual encoders: BGE-m3~\cite{bge-m3} and two Sentence-BERT variants~\cite{reimers-2019-sentence-bert}, \texttt{paraphrase-multilingual-MiniLM-L12-v2} and \texttt{paraphrase-multilingual-mpnet-base-v2}. We denote them as \texttt{p-mMiniLM} and \texttt{p-mMpNet} for compactness in tables.


\newcolumntype{R}{>{\raggedleft\arraybackslash}X}

\begin{table}[!htbp]
\centering
\footnotesize
\setlength{\tabcolsep}{3pt}
\renewcommand{\arraystretch}{1.08}

\begin{tabularx}{\linewidth}{@{}l *{6}{R}@{}}
\toprule
& \multicolumn{2}{c}{\textbf{\shortstack{$p_{\mathrm{ret}}$}}}
& \multicolumn{2}{c}{\textbf{\shortstack{$p_{\mathrm{gold}}$}}}
& \multicolumn{2}{c}{\textbf{\shortstack{$p_{\mathrm{cult}}$}}} \\
\cmidrule(lr){2-3}\cmidrule(lr){4-5}\cmidrule(lr){6-7}
\textbf{Encoder}
& \textbf{MLRS} & \textbf{DeLP}
& \textbf{MLRS} & \textbf{DeLP}
& \textbf{MLRS} & \textbf{DeLP} \\
\midrule
\texttt{bge-m3}    & 0.994 & 0.142 & 0.914 & 0.336 & 0.916 & 0.335 \\
\texttt{p-mMiniLM} & 0.997 & 0.145 & 0.915 & 0.321 & 0.917 & 0.320 \\
\texttt{p-mMpNet}  & 0.996 & 0.131 & 0.917 & 0.311 & 0.920 & 0.310 \\
\bottomrule
\end{tabularx}
\vspace{-2mm}
\caption{Pearson’s $r$ between preference and priors for before (MLRS) and after (DeLP) calibration.} 
\label{tab:corr_preference_priors}
\vspace{-3mm}
\end{table}

\paragraph{Emergence of Monolingual Alignment.}


After calibration, we find that the preference landscape shifts qualitatively from the raw preference as in Table~\ref{tab:delp}. The previously dominant English preference largely disappears, and the strongest signal consistently moves to the diagonal ($L_q = L_d$). This reveals that 
retrievers fundamentally favor monolingual alignment—the matching of query and document in the same language. We also observe 
that queries favor the linguistically or regionally related languages, such as Korean with Japanese. Overall, the DeLP score suggests that much of the apparent English preference in prior protocols was induced by structural priors, while the residual preference signal is dominated by query-language alignment and interpretable related-language effects. 
For a more detailed DeLP score, refer to Appendix~\ref{app:delp_details}.

\paragraph{Correlation Analysis.} 


To validate DeLP, we compute the correlation between preference scores and priors before and after calibration as shown in Table~\ref{tab:corr_preference_priors}. Raw scores (MLRS) are highly correlated with all three priors (exposure, gold-availability, and cultural), suggesting that existing language-preference measurements largely reflect prior-driven preference rather than intrinsic model preference.
After calibration, these correlations drop sharply after applying DeLP. This confirms that DeLP effectively decouples intrinsic model tendencies from the structural signals.

\begin{figure*}[t]

  \centering
  \includegraphics[width=0.93\textwidth]{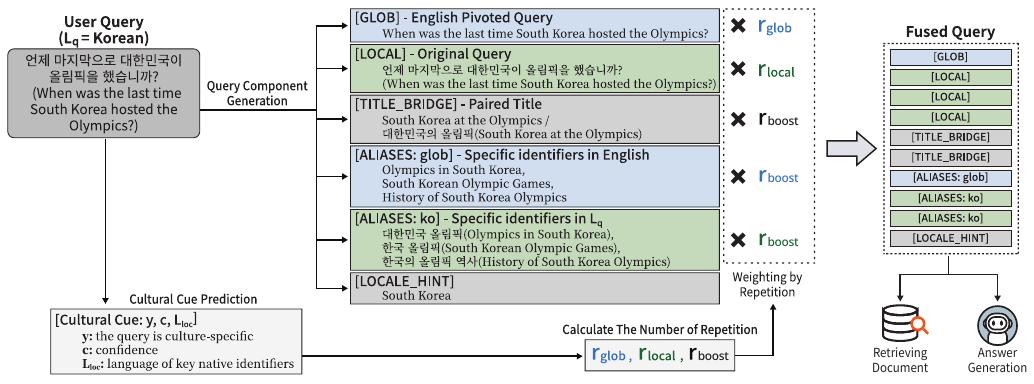}
  \vspace{-3mm}
  \caption{Overview of DELTA query fusion. DELTA fuses global and local query segments into a single preference-aligned query using lightweight repetition-based weighting.}
  \label{fig:delta}
  \vspace{-4mm}
\end{figure*}

\section{Debiasing mRAG through Preference-Aligned Augmentation}

Monolingual alignment in Section~\ref{sec:calibration} reveals that \textit{retrievers intrinsically perform best} when the \textit{query language matches the document language}. This suggests that while English pivoting provides coverage (due to gold availability), it often sacrifices the retrieval anchors present in the user's native tongue. Motivated by this point, we propose DELTA (\textbf{DE}biased \textbf{L}anguage preference guided \textbf{T}ext \textbf{A}ugmentation), a lightweight query reformulation strategy that injects preference-aligned cues into a single fused query.

\subsection{Query Fusion with Native Anchors}
DELTA aims to maximize the benefits of both global coverage and local discriminative matching. As illustrated in Figure~\ref{fig:delta}, given a local question $q_{\text{local}}$, DELTA constructs an English pivot $q_{\text{glob}}$ and extracts a set of cultural identifiers—canonical titles $(t_{\text{glob}}, t_{\text{loc}})$, aliases, and a regional hint—using a frozen LLM (Instruction in Appendix~\ref{app:prompt}). These elements are then concatenated into a single query $Q_{\text{fused}}$ composed of five 
segments, each optionally weighted by cultural cues' confidence score:

\begin{itemize}[leftmargin=*, nosep]
\item \texttt{[GLOB]}: The English pivot $q_{\text{glob}}$.
\item \texttt{[LOCAL]}: The original query $q_{\text{local}}$ to leverage monolingual alignment.
\item \texttt{[TITLE\_BRIDGE]}: Paired titles
$(t_{\text{glob}}, t_{\text{loc}})$ to facilitate cross-lingual mapping.
\item \texttt{[ALIASES]} and \texttt{[LOCALE\_HINT]}: Specific identifiers that serve as stable retrieval anchors.
\end{itemize}

\subsection{Repetition-based Weighting}

To implement the debiased preference control, DELTA utilizes a repetition-based weighting policy\cite{wang2023query2doc}.
We first predict a cultural cue $(y,c,L_{\mathrm{loc}})$, where $y\in\{0,1\}$ indicates whether the query is culture-specific, $c\in[0,1]$ is the confidence score, and $L_{\mathrm{loc}}$ is the language of the key native identifiers. 
This cue determines how strongly we upweight locale-specific blocks versus the global English back-off when forming the fused query $Q_{\text{fused}}$. 
We map the confidence score $c$ into three discrete repetition levels using two thresholds $\tau_{\mathrm{low}}$ and $\tau_{\mathrm{high}}$.
We then set repetition counts for the local block \texttt{[LOCAL:$L_{\mathrm{loc}}$]} and the global pivot block \texttt{[GLOB]} as:

\vspace{-3mm}
\par\begingroup\small
\begin{equation}
\label{eq:delta_repetition_glob}
\begin{aligned}
r_{\text{local}} &=
\begin{cases}
1 + \mathbb{I}[c\ge\tau_{\mathrm{low}}] + \mathbb{I}[c\ge\tau_{\mathrm{high}}], & y=1\\
1, & y=0
\end{cases}
\\[2pt]
r_{\text{glob}} &=
\begin{cases}
1 + \mathbb{I}[c<\tau_{\mathrm{low}}], & y=1\\
2, & y=0
\end{cases}
\end{aligned}
\end{equation}
\endgroup\par
\vspace{-1mm}

$c<\tau_{\mathrm{low}}$ triggers no additional upweighting, $\tau_{\mathrm{low}}\le c<\tau_{\mathrm{high}}$ adds one extra repetition, and $c\ge\tau_{\mathrm{high}}$ adds two, yielding $r_{\text{local}}\in\{1,2,3\}$ and $r_{\text{glob}}\in\{1,2\}$ . Intuitively, we upweight high-confidence culture-specific queries toward the local expression to preserve culturally grounded identifiers, while non-culture-specific queries mildly favor the global pivot for robust back-off. 
In addition, when $y{=}1$ and $c\ge\tau_{\mathrm{boost}}$, we duplicate local-side disambiguation anchors (i.e., \texttt{[TITLE\_BRIDGE]} and \texttt{[ALIASES]}) once more to further emphasize native surface-form anchors and reduce entity ambiguity.
Overall, DELTA realizes preference control via text-only weighting, and all concrete hyperparameter values are reported in Appendix~\ref{app:delta_repetition}.

\section{Experiments}
\subsection{Experimental Setup}
\textbf{Baselines.}
(1) \textbf{MultiRAG}: Retrieve and re-rank from multilingual datastores using the original MKQA query, then generate the answer in the query language.~\cite{chirkova-etal-2024-retrieval}
(2) \textbf{CrossRAG}: Run the same multilingual retrieval as MultiRAG, translate the retrieved passages into a single pivot language (English).~\cite{ranaldi2025multilingual} 
(3) \textbf{DKM-RAG}: Translate the retrieved passages into the query language, use an LLM to produce multiple refined passages.~\cite{park2025investigating} 
(4) \textbf{QTT-RAG}: Translate retrieved passages and attach translation-quality tags so the generator can decide which contexts to trust.~\cite{moon2025quality} 
(5) \textbf{English Translation}: Translate the original query into the global pivot language.

\begin{table*}[t]
\centering
\footnotesize
\setlength{\tabcolsep}{4.0pt}
\renewcommand{\arraystretch}{1.02}

\begin{tabular*}{\textwidth}{@{\extracolsep{\fill}}lcccccccccc}
\toprule
\textbf{Method} & \textbf{\textit{en}} & \textbf{\textit{ar}} & \textbf{\textit{es}} & \textbf{\textit{zh}} &
\textbf{\textit{ja}} & \textbf{\textit{de}} & \textbf{\textit{ko}} & \textbf{\textit{th}} &
\textbf{AVG $\uparrow$} & \textbf{Latency $\downarrow$} \\
\midrule

\rowcolor{llmhead}\multicolumn{11}{l}{\textbf{Qwen3-235b-a22b-2507}}\\
\rowcolor{llmshade}\multicolumn{11}{c}{\textbf{\textit{Document Level}}}\\
MultiRAG & \second{70.05} & 47.79 & \best{63.76} & 37.52 & 46.60 & \best{63.81} & 40.14 & 40.73 & 51.30 & 1.38 \\
CrossRAG & 68.21 & 43.95 & 61.14 & 37.81 & 44.75 & 60.16 & 38.13 & 42.87 & 49.63 & 1.29 \\
DKM-RAG  & 69.13 & 42.69 & 62.12 & 35.13 & 43.90 & 61.13 & 39.49 & 38.88 & 49.06 & 3.80 \\
QTT-RAG  & \best{70.11} & 46.44 & 63.02 & 37.68 & 46.94 & 62.79 & 44.13 & 42.12 & 51.65 & 1.80 \\
\rowcolor{llmshade}\multicolumn{11}{c}{\textbf{\textit{Query Level}}}\\
English Translation & - & \second{55.14} & 61.94 & \second{59.53} & \second{59.29} & 60.72 & \second{54.57} & \second{60.46} & \second{58.81} & \second{1.17} \\
DELTA (ours) & 63.85 & \best{62.55} & \second{63.03} & \best{62.59} & \best{62.38} & \second{62.86} & \best{63.26} & \best{62.51} & \best{62.88} & \best{1.13} \\
\midrule

\rowcolor{llmhead}\multicolumn{11}{l}{\textbf{Gemini-2.5-flash}}\\
\rowcolor{llmshade}\multicolumn{11}{c}{\textbf{\textit{Document Level}}}\\
MultiRAG & 58.26 & 40.79 & 55.11 & 30.81 & 44.26 & 53.52 & 35.97 & 31.65 & 43.80 & \second{1.53} \\
CrossRAG & 63.40 & 41.87 & 57.24 & 29.74 & 44.14 & \second{56.80} & 36.09 & 32.49 & 45.22 & 2.60 \\
DKM-RAG  & \second{64.21} & 39.41 & \best{59.26} & 31.34 & 43.45 & \best{57.74} & 37.26 & 33.64 & 45.79 & 5.63 \\
QTT-RAG  & \best{65.32} & 42.64 & \second{57.81} & 31.56 & 45.18 & 56.27 & 40.65 & 35.97 & 46.93 & 5.55 \\
\rowcolor{llmshade}\multicolumn{11}{c}{\textbf{\textit{Query Level}}}\\
English Translation & - & \second{48.44} & 55.84 & \second{53.59} & \second{53.68} & 55.17 & \second{47.67} & \second{54.86} & \second{52.75} & 1.55 \\
DELTA (ours) & 56.97 & \best{56.45} & 55.95 & \best{55.83} & \best{56.18} & 55.98 & \best{56.44} & \best{56.45} & \best{56.28} & \best{1.48} \\
\midrule

\rowcolor{llmhead}\multicolumn{11}{l}{\textbf{Deepseek-chat-v3.1}}\\
\rowcolor{llmshade}\multicolumn{11}{c}{\textbf{\textit{Document Level}}}\\
MultiRAG  & 60.77 & 43.64 & 56.22 & 33.14 & 44.72 & 54.16 & 34.21 & 36.80 & 45.46 & 2.56 \\
CrossRAG & 67.83 & 48.34 & \second{62.24} & 39.05 & 49.27 & \second{61.33} & 39.85 & 45.70 & 51.70 & 2.64 \\
DKM-RAG  & \second{67.84} & 44.07 & \best{62.49} & 37.63 & 45.66 & \best{61.65} & 40.30 & 40.38 & 50.00 & 2.39 \\
QTT-RAG  & \best{68.28} & 46.13 & 61.81 & 37.24 & 47.36 & 60.48 & 41.06 & 41.29 & 50.46 & \second{1.93} \\
\rowcolor{llmshade}\multicolumn{11}{c}{\textbf{\textit{Query Level}}}\\
English Translation & - & \second{50.97} & 58.32 & \second{56.11} & \second{56.52} & 56.92 & \second{50.49} & \best{57.52} & \second{55.26} & 2.05 \\
DELTA (ours) & 59.85 & \best{59.46} & 58.61 & \best{59.67} & \best{59.02} & 59.25 & \best{53.51} & \second{56.45} & \best{58.23} & \best{1.13} \\
\bottomrule
\end{tabular*}
\vspace{-2mm}
\caption{Main results (end-to-end mRAG performance). We use bge-m3~\cite{bge-m3} for retrieval, and evaluate with character 3-gram recall~\cite{chirkova-etal-2024-retrieval}. \best{Best}, \second{second-best} AVG (mean) are computed per generator and language. We report generation time as latency.} 
\label{tab:main_results}
\vspace{-5mm}
\end{table*}

\vspace{-2mm}
\subsection{Results and Analysis}
\paragraph{Main Results.}
Table~\ref{tab:main_results} shows that DELTA achieves the best average performance for each generator and is comparable to, or better than, document-level frameworks that require substantially higher cost due to the document's long context length. The gains are particularly pronounced on non-English queries, indicating that preference-aligned query augmentation is more effective than relying on document-side transformations in multilingual settings.
DELTA provides little benefit on English queries (the \texttt{en} column in Table~\ref{tab:main_results}) because the local query and the global pivot become nearly identical when $L_q=\texttt{en}$. As a result, DELTA injects redundant segments with repeated, overlapping content, which unnecessarily lengthens the query and can dilute the useful signal for retrieval, resulting in no gain or even a slight degradation.
\vspace{-2mm}

\begin{table}[!hbp]
\centering
\footnotesize 
\setlength{\tabcolsep}{12pt} 
\renewcommand{\arraystretch}{0.92} 

\begin{tabular}{l r} 
\toprule
\textbf{Statistic} & \textbf{Value} \\
\midrule
Queries ($N$) & 16,828 \\
Newly recovered queries ($N_{\text{new}}$) & 1,235 \\
\midrule
Gold best-rank (mean) & 10.39 \\
Gold best-rank (median) & 5 \\
Top-10 rate & 66.23\% \\
Rank in [10,49] rate & 35.14\% \\
Rank in [40,50] rate & 4.62\% \\
\bottomrule
\end{tabular}
\vspace{-2mm}
\caption{Retrieval rank analysis, which reports the rank distribution of gold passages newly recovered by DELTA relative to English-pivot.}
\vspace{-6mm}
\label{tab:delta_ret_newhit}
\end{table}

\paragraph{Analyzing Gold Passage Recall and Ranking.}
To investigate how DELTA recovers missing gold evidence to improve the overall mRAG system, we compare its retrieval performance against English-pivot retrieval across seven languages (ar, de, es, ja, ko, th, zh), totaling 16,828 queries. As shown in Table~\ref{tab:delta_ret_newhit}, while English pivoting provides coverage due to English-heavy gold availability, it often degrades native surface-form anchors—such as titles, aliases, and original scripts—that are critical for precise entity matching. DELTA restores these anchors, facilitating better alignment with English gold documents. Among 1,235 newly recovered queries, DELTA achieves a mean best rank of 10.39 (median 5) with a 66.2\% Top-10 entry rate. This indicates that DELTA does not merely rescue missed gold pages near the cutoff but significantly elevates them to high-ranking, actionable positions.

\begin{table}[!h]
\centering
\small
\setlength{\tabcolsep}{3pt}
\renewcommand{\arraystretch}{1.08}

\resizebox{\columnwidth}{!}{\begin{tabular}{lcccccccc}
\toprule
\textbf{Method} & \textbf{ar} & \textbf{de} & \textbf{es} & \textbf{ja} & \textbf{ko} & \textbf{th} & \textbf{zh} & \textbf{Avg} \\
\midrule
Orig
& 63.68 & 62.46 & 63.26 & 63.26 & 62.84 & 63.37 & 63.04 & 63.13 \\
+Global
& \second{71.42} & \second{72.02} & \second{71.37} & \second{71.51} & \second{71.77} & \second{71.70} & \second{71.57} & \second{71.62} \\
+Title
& 68.17 & 67.83 & 67.75 & 67.78 & 67.93 & 68.17 & 68.34 & 68.00 \\
+Aliases
& 68.14 & 67.46 & 67.43 & 67.38 & 68.61 & 68.15 & 68.13 & 67.90 \\
+Locale
& 67.57 & 67.63 & 67.78 & 67.89 & 67.81 & 67.65 & 67.44 & 67.68 \\
All cues
& \best{72.99} & \best{73.01} & \best{72.48} & \best{73.26} & \best{72.88} & \best{72.93} & \best{72.70} & \best{72.89} \\
\bottomrule
\end{tabular}
}
\vspace{-2mm}
\caption{Cue ablations for DELTA with fixed evidence. We incrementally add query-side cues and report end-to-end generation accuracy.}
\label{tab:delta_cue_ablation}
\end{table}

\paragraph{Impact of Cues on Evidence Interpretation.}
To isolate generation-stage effects from retrieval-stage effects, we conduct a cue ablation study under a fixed-evidence setting, which is reported in Table~\ref{tab:delta_cue_ablation}.  Specifically, we first retrieve passages and re-rank, then hold those retrieved passages constant while modifying only the query-side cues at generation time. This design ensures that any observed performance differences stem solely from how the generator interprets the fixed evidence under different query formulations, not from changes in retrieved content.
Under this setup, the global pivot cue significantly outperforms the original query, indicating that concise global English paraphrasing aids the generator in aligning evidence. Bridge cues also provide independent gains, showing that even when the retrieved evidence context is held fixed, varying only the query-side cues at generation time improves the model's ability to select precise evidence spans. The best performance across all languages is achieved by combining all cues, suggesting that bridge cues offer critical disambiguation and entity grounding.

\paragraph{Latency Analysis.}
To assess the efficiency of DELTA, we report average end-to-end latency (wall-clock time in seconds per query, averaged over all test queries) in the rightmost column of Table~\ref{tab:main_results}. We provide detailed per-language latency measurements in Appendix~\ref{app:latency}.
DELTA maintains high efficiency by generating a single fused query and avoiding document translation. 
It can even be faster than English Translation; by incorporating local cues and disambiguation anchors, DELTA enables direct retrieval, reducing the overhead of processing overly generic English-only signals.


\section{Related Works}

\subsection{Multilingual RAG}
Prior work in mRAG has explored how performance varies with the query language~\cite{ranaldi2025multilingual, longpre-etal-2021-mkqa}, the language of relevant or irrelevant evidence~\cite{qi-etal-2025-consistency, wu2024languagesequalinsightsmultilingual}, as well as document ordering and prompting strategies that affect how models consume multilingual contexts~\cite{sharma2024fauxpolyglotstudyinformation, wu2024languagesequalinsightsmultilingual, shankar-etal-2024-context, ki2025linguistic}. A common and effective heuristic is pivot translation, where non-English queries are translated into English before retrieval, often producing large gains~\cite{asai2021xor, ranaldi2025multilingual}.
However, much of the existing analysis of why pivot translation helps centers on the generation stage (e.g., English-centric generation competence, translation noise, and cross-lingual drift), which motivates generator-side interventions such as translation-aware prompting or decoding-time control~\cite{sharma2024fauxpolyglotstudyinformation, moon2025quality}. In contrast, our work focuses on a retrieval-side explanation: we empirically show that gold evidence is structurally skewed toward English corpora.

\subsection{Language Preference}
In mRAG, language preference is shown both in retrieval (over-retrieving high-resource languages) and in generation (differentially using evidence by language even under \textit{matched relevance}---a setting where the gold passage or core supporting evidence is correctly retrieved across all compared conditions), degrading consistency and downstream quality~\cite{park2025investigating}.
Existing measurements of language preference in mRAG commonly rely on behavioral proxies, such as comparing outputs across query languages via information overlap~\cite{sharma2024fauxpolyglotstudyinformation} or embedding similarity to references~\cite{park2025investigating}, and, in more controlled settings, analyzing citation or attribution behavior as evidence that language varies while other variables are fixed~\cite{ki2025linguistic, qi-etal-2025-consistency}. While prior approaches offer useful signals, they miss a key confound in mRAG: structural priors can dominate preference scores. We therefore debias preference by regressing out these priors and using the residual as the preference signal.

\section{Conclusion}
We demonstrate that gains from English pivoting in mRAG stem from retrieval-side evidence imbalance, which biases preference measurements. We address this with DeLP, a debiased metric that calibrates structural priors to reveal preference shifts toward the query language. Leveraging DeLP, we introduce DELTA, a lightweight query reformulation strategy that fuses global and local cues into a single query, consistently outperforming baselines.

\section*{Limitations}
First, our debiasing targets retriever-level preference, while generator-level preference can still remain. Therefore, extending debiasing to how generators consume multilingual evidence is an important direction for future work. Second, our conclusions are drawn from a Wikipedia-based mRAG setup. 
Evaluating DeLP and DELTA on broader, domain-specific multilingual corpora is therefore necessary to assess their generalizability.
Third, DELTA controls the balance between global and local signals using simple repetition, which is coarse. More precise and principled weighting or adaptive control logic could further improve effectiveness and stability.

\section*{Ethics Statement}
We conduct our experiments using publicly available multilingual datasets, knowledge sources, and models that are widely used in the research community and released under established data-sharing and licensing guidelines. We follow the usage protocols and license agreements specified by the original providers. While these resources are designed to reduce harmful biases and inappropriate content, they may still contain artifacts of data imbalance and may not fully represent the diversity of languages, dialects, and cultural contexts. Our work analyzes and mitigates retrieval-side evidence imbalance and does not involve human subject data, user interaction logs, or the collection of personally identifiable information. We encourage future deployments to consider downstream risks such as uneven coverage across languages and potential disparities in answer quality for under-resourced communities.

\section*{Acknowledgments}
This work was supported by the Institute of Information \& Communications Technology Planning \& Evaluation (IITP) grant funded by the Korea government (MSIT) [RS-2021-II211341, Artificial Intelligence Graduate School Program (Chung-Ang University)] and the National Research Foundation of Korea(NRF) grant funded by the Korea government(MSIT) (RS-2026-25494299).
This research was supported by the Chung-Ang University Graduate Research Scholarship in 2025.

\bibliography{main}

@inproceedings{wang2023query2doc,
  title={Query2doc: Query Expansion with Large Language Models},
  author={Wang, Liang and Yang, Nan and Wei, Furu},
  booktitle={Proceedings of the 2023 Conference on Empirical Methods in Natural Language Processing},
  pages={9414--9423},
  year={2023}
}

@article{hurst2024gpt,
  title={Gpt-4o system card},
  author={Hurst, Aaron and Lerer, Adam and Goucher, Adam P and Perelman, Adam and Ramesh, Aditya and Clark, Aidan and Ostrow, AJ and Welihinda, Akila and Hayes, Alan and Radford, Alec and others},
  journal={arXiv preprint arXiv:2410.21276},
  year={2024}
}

@misc{sharma2024fauxpolyglotstudyinformation,
      title={Faux Polyglot: A Study on Information Disparity in Multilingual Large Language Models}, 
      author={Nikhil Sharma and Kenton Murray and Ziang Xiao},
      year={2024},
      eprint={2407.05502},
      archivePrefix={arXiv},
      primaryClass={cs.CL},
      url={https://arxiv.org/abs/2407.05502}, 
}

@article{ki2025linguistic,
  title={Linguistic Nepotism: Trading-off Quality for Language Preference in Multilingual RAG},
  author={Ki, Dayeon and Carpuat, Marine and McNamee, Paul and Khashabi, Daniel and Yang, Eugene and Lawrie, Dawn and Duh, Kevin},
  journal={arXiv preprint arXiv:2509.13930},
  year={2025}
}

@inproceedings{shankar-etal-2024-context,
    title = "In-context Mixing ({ICM}): Code-mixed Prompts for Multilingual {LLM}s",
    author = "Shankar, Bhavani  and
      Jyothi, Preethi  and
      Bhattacharyya, Pushpak",
    editor = "Ku, Lun-Wei  and
      Martins, Andre  and
      Srikumar, Vivek",
    booktitle = "Proceedings of the 62nd Annual Meeting of the Association for Computational Linguistics (Volume 1: Long Papers)",
    month = aug,
    year = "2024",
    address = "Bangkok, Thailand",
    publisher = "Association for Computational Linguistics",
    url = "https://aclanthology.org/2024.acl-long.228/",
    doi = "10.18653/v1/2024.acl-long.228",
    pages = "4162--4176"
}

@inproceedings{asai2021xor,
  title={XOR QA: Cross-lingual open-retrieval question answering},
  author={Asai, Akari and Kasai, Jungo and Clark, Jonathan H and Lee, Kenton and Choi, Eunsol and Hajishirzi, Hannaneh},
  booktitle={Proceedings of the 2021 conference of the North American chapter of the association for computational linguistics: human language technologies},
  pages={547--564},
  year={2021}
}

@misc{wu2024languagesequalinsightsmultilingual,
      title={Not All Languages are Equal: Insights into Multilingual Retrieval-Augmented Generation}, 
      author={Suhang Wu and Jialong Tang and Baosong Yang and Ante Wang and Kaidi Jia and Jiawei Yu and Junfeng Yao and Jinsong Su},
      year={2024},
      eprint={2410.21970},
      archivePrefix={arXiv},
      primaryClass={cs.CL},
      url={https://arxiv.org/abs/2410.21970}, 
}

@article{ranaldi2025multilingual,
  title={Multilingual Retrieval-Augmented Generation for Knowledge-Intensive Task},
  author={Ranaldi, Leonardo and Haddow, Barry and Birch, Alexandra},
  journal={arXiv preprint arXiv:2504.03616},
  year={2025}
}

@article{li2025language, 
  title={Language Drift in Multilingual Retrieval-Augmented Generation: Characterization and Decoding-Time Mitigation},
  author={Li, Bo and Xu, Zhenghua and Xie, Rui},
  journal={arXiv preprint arXiv:2511.09984},
  year={2025}
}

@inproceedings{moon2025quality,
  title={Quality-Aware Translation Tagging in Multilingual RAG system},
  author={Moon, Hoyeon and Kim, Byeolhee and Verma, Nikhil},
  booktitle={Proceedings of the 5th Workshop on Multilingual Representation Learning (MRL 2025)},
  pages={161--177},
  year={2025}
}

@inproceedings{chirkova-etal-2024-retrieval,
    title = "Retrieval-augmented generation in multilingual settings",
    author = "Chirkova, Nadezhda  and
      Rau, David  and
      D{\'e}jean, Herv{\'e}  and
      Formal, Thibault  and
      Clinchant, St{\'e}phane  and
      Nikoulina, Vassilina",
    editor = "Li, Sha  and
      Li, Manling  and
      Zhang, Michael JQ  and
      Choi, Eunsol  and
      Geva, Mor  and
      Hase, Peter  and
      Ji, Heng",
    booktitle = "Proceedings of the 1st Workshop on Towards Knowledgeable Language Models (KnowLLM 2024)",
    month = aug,
    year = "2024",
    address = "Bangkok, Thailand",
    publisher = "Association for Computational Linguistics",
    url = "https://aclanthology.org/2024.knowllm-1.15/",
    doi = "10.18653/v1/2024.knowllm-1.15",
    pages = "177--188"
}

@inproceedings{park2025investigating,
    title = "Investigating Language Preference of Multilingual {RAG} Systems",
    author = "Park, Jeonghyun  and
      Lee, Hwanhee",
    editor = "Che, Wanxiang  and
      Nabende, Joyce  and
      Shutova, Ekaterina  and
      Pilehvar, Mohammad Taher",
    booktitle = "Findings of the Association for Computational Linguistics: ACL 2025",
    month = jul,
    year = "2025",
    address = "Vienna, Austria",
    publisher = "Association for Computational Linguistics",
    url = "https://aclanthology.org/2025.findings-acl.295/",
    doi = "10.18653/v1/2025.findings-acl.295",
    pages = "5647--5675",
    ISBN = "979-8-89176-256-5"
}

@inproceedings{qi-etal-2025-consistency,
    title = "On the Consistency of Multilingual Context Utilization in Retrieval-Augmented Generation",
    author = "Qi, Jirui  and
      Fern{\'a}ndez, Raquel  and
      Bisazza, Arianna",
    editor = "Adelani, David Ifeoluwa  and
      Arnett, Catherine  and
      Ataman, Duygu  and
      Chang, Tyler A.  and
      Gonen, Hila  and
      Raja, Rahul  and
      Schmidt, Fabian  and
      Stap, David  and
      Wang, Jiayi",
    booktitle = "Proceedings of the 5th Workshop on Multilingual Representation Learning (MRL 2025)",
    month = nov,
    year = "2025",
    address = "Suzhuo, China",
    publisher = "Association for Computational Linguistics",
    url = "https://aclanthology.org/2025.mrl-main.15/",
    doi = "10.18653/v1/2025.mrl-main.15",
    pages = "199--225",
    ISBN = "979-8-89176-345-6"
}

@inproceedings{lewis2020retrieval,
author = {Lewis, Patrick and Perez, Ethan and Piktus, Aleksandra and Petroni, Fabio and Karpukhin, Vladimir and Goyal, Naman and K\"{u}ttler, Heinrich and Lewis, Mike and Yih, Wen-tau and Rockt\"{a}schel, Tim and Riedel, Sebastian and Kiela, Douwe},
title = {Retrieval-augmented generation for knowledge-intensive NLP tasks},
year = {2020},
isbn = {9781713829546},
publisher = {Curran Associates Inc.},
address = {Red Hook, NY, USA},
booktitle = {Proceedings of the 34th International Conference on Neural Information Processing Systems},
articleno = {793},
numpages = {16},
location = {Vancouver, BC, Canada},
series = {NIPS '20}
}

@misc{bge-m3,
      title={BGE M3-Embedding: Multi-Lingual, Multi-Functionality, Multi-Granularity Text Embeddings Through Self-Knowledge Distillation}, 
      author={Jianlv Chen and Shitao Xiao and Peitian Zhang and Kun Luo and Defu Lian and Zheng Liu},
      year={2024},
      eprint={2402.03216},
      archivePrefix={arXiv},
      primaryClass={cs.CL}
}

@inproceedings{reimers-2019-sentence-bert,
    title = "Sentence-BERT: Sentence Embeddings using Siamese BERT-Networks",
    author = "Reimers, Nils and Gurevych, Iryna",
    booktitle = "Proceedings of the 2019 Conference on Empirical Methods in Natural Language Processing",
    month = "11",
    year = "2019",
    publisher = "Association for Computational Linguistics",
    url = "http://arxiv.org/abs/1908.10084",
}

@article{longpre-etal-2021-mkqa,
    title = "{MKQA}: A Linguistically Diverse Benchmark for Multilingual Open Domain Question Answering",
    author = "Longpre, Shayne  and
      Lu, Yi  and
      Daiber, Joachim",
    editor = "Roark, Brian  and
      Nenkova, Ani",
    journal = "Transactions of the Association for Computational Linguistics",
    volume = "9",
    year = "2021",
    address = "Cambridge, MA",
    publisher = "MIT Press",
    url = "https://aclanthology.org/2021.tacl-1.82/",
    doi = "10.1162/tacl_a_00433",
    pages = "1389--1406"
}

@article{yang2025qwen3,
  title={Qwen3 technical report},
  author={Yang, An and Li, Anfeng and Yang, Baosong and Zhang, Beichen and Hui, Binyuan and Zheng, Bo and Yu, Bowen and Gao, Chang and Huang, Chengen and Lv, Chenxu and others},
  journal={arXiv preprint arXiv:2505.09388},
  year={2025}
}

@article{comanici2025gemini,
  title={Gemini 2.5: Pushing the frontier with advanced reasoning, multimodality, long context, and next generation agentic capabilities},
  author={Comanici, Gheorghe and Bieber, Eric and Schaekermann, Mike and Pasupat, Ice and Sachdeva, Noveen and Dhillon, Inderjit and Blistein, Marcel and Ram, Ori and Zhang, Dan and Rosen, Evan and others},
  journal={arXiv preprint arXiv:2507.06261},
  year={2025}
}

@article{liu2024deepseek,
  title={Deepseek-v3 technical report},
  author={Liu, Aixin and Feng, Bei and Xue, Bing and Wang, Bingxuan and Wu, Bochao and Lu, Chengda and Zhao, Chenggang and Deng, Chengqi and Zhang, Chenyu and Ruan, Chong and others},
  journal={arXiv preprint arXiv:2412.19437},
  year={2024}
}

@inproceedings{rau2024bergen,
  title={Bergen: A benchmarking library for retrieval-augmented generation},
  author={Rau, David and D{\'e}jean, Herv{\'e} and Chirkova, Nadezhda and Formal, Thibault and Wang, Shuai and Clinchant, St{\'e}phane and Nikoulina, Vassilina},
  booktitle={Findings of the Association for Computational Linguistics: EMNLP 2024},
  pages={7640--7663},
  year={2024}
}

@article{zhang2023don,
  title={Don't trust ChatGPT when your question is not in English: a study of multilingual abilities and types of LLMs},
  author={Zhang, Xiang and Li, Senyu and Hauer, Bradley and Shi, Ning and Kondrak, Grzegorz},
  journal={arXiv preprint arXiv:2305.16339},
  year={2023}
}

\clearpage
\appendix
\section*{Appendix}

\section{The Use of Large Language Models}
We write the manuscript ourselves, and an LLM (ChatGPT-5.2) is used solely for refinement—style, clarity, and grammar. It is not used for ideation or content generation.

\section{Implementation Details}
\label{app:detail}
We adopt the multilingual retrieval baseline of Bergen~\cite{chirkova-etal-2024-retrieval}, which retrieves evidence from a datastore spanning all languages. For generations, we follow Bergen’s prompting setup and use the basic\_translated\_langspec template (Figure~\ref{fig:prompt_templates}) to produce the final mRAG response. Building on this standardized pipeline, we conduct a series of experiments to quantify language preference in mRAG under the Bergen framework, which systematically examines the key components and practical adjustments necessary for a robust multilingual RAG baseline. We use a robust multilingual LLM, qwen3-235b-a22b-2507, to translate the user's local question into the global pivot language, English. We instruct GPT-4o-mini~\cite{hurst2024gpt} to generate a compact lexical bundle that supplies candidate titles, aliases, and a short disambiguation hint for the global and local segments. All LLM calls are made using the OpenRouter API. We conduct our experiments using an AMD EPYC 7313 CPU (3.0 GHz) paired with four NVIDIA RTX 6000 Ada GPUs. We use Python 3.11.5 and PyTorch 2.3.1 for the software environment.

\section{Language Notation}
We use standard ISO 639-1 language codes to denote the languages in our experiments. Specifically, en denotes English, ar represents Arabic, es corresponds to Spanish, zh refers to Chinese (Simplified), ja indicates Japanese, de stands for German, ko denotes Korean, and th denotes Thai. These concise codes facilitate consistent identification and processing of language-specific data across datasets and models in multilingual NLP research.

\section{Dataset Details \& Statistics}
\label{app:statistics}

Wikipedia is a widely adopted knowledge source in both monolingual RAG and mRAG systems, as it provides broad topical coverage and is commonly used to benchmark RAG pipelines. In most experiments, we retrieve from various linguistic data sources from (i) the KILT snapshot of English Wikipedia\footnote{\url{https://huggingface.co/datasets/facebook/kilt_wikipedia}} and (ii) the Wikipedia edition in the user’s local language\footnote{\url{https://huggingface.co/datasets/wikimedia/wikipedia}}. This two-source design reflects a standard and practical mRAG setting where English serves as a high-coverage reference corpus while local-language Wikipedia captures language-specific evidence and terminology.

We report summary statistics for the data resources used in our experiments in Table~\ref{tab:data_statistics}. MKQA is our primary evaluation dataset, and we provide the number of examples along with the median lengths of questions and answers. We also use Wikipedia as the external corpus for the retriever datastore; its statistics, including the number of passages and their median lengths, are likewise presented in Table~\ref{tab:data_statistics}. These statistics provide an overview of the datasets and corpora underlying our experimental setup.

\section{Raw Language Preference Score}
\paragraph{MLRS.}
Following the standard MultiLingualRankShift (MLRS) protocol, we quantify retriever-level language preference by measuring how much the ranks of non-query-language documents improve after being translated into the query language~\cite{park2025investigating}.
For each query $q$ with language $L_q$, we first retrieve a ranked list $D_q$ from a multilingual datastore, assigning each document $d\in D_q$ an initial rank $r^{\text{init}}_d$.
We then translate documents with $L_d\neq L_q$ into $L_q$ and re-rank them using the same retriever, obtaining $r^{\text{re-rank}}_d$.
The (non-negative) rank gain is computed as $\Delta r_d=\max\!\bigl(r^{\text{init}}_d-r^{\text{re-rank}}_d,\,0\bigr)$, and aggregated per query as $\Delta r_q=\sum_{d}\Delta r_d$.
Normalizing by the maximum possible gain $\Delta r^{\max}_q=\sum_{d}(r^{\text{init}}_d-1)$ yields the query-level score
$\mathrm{MLRS}_q=\frac{\Delta r_q}{\Delta r^{\max}_q}\times 100$ (or $0$ if $\Delta r^{\max}_q=0$),
and the final MLRS is the average over queries.

\paragraph{Results.} Table~\ref{tab:mlrs} reports the retriever’s language preference scores before calibration with DeLP. Overall, we observe three consistent patterns. First, cross-lingual retrieval ($L_q \neq L_d$) generally yields lower MLRS than monolingual retrieval, indicating that cross-lingual matching is less preferred in most cases. Second, English emerges as a dominant target language: when the retrieved document language $L_d$ is English, the retriever attains near-maximum preference scores and often even surpasses monolingual settings, consistent with English-heavy pretraining and stronger English representations. Third, cross-lingual preference is partly modulated by linguistic relatedness: closely related Romance languages (fr/it/pt/es) preserve relatively high cross-lingual scores, while East Asian pairs (ko/ja/zh) show moderate but noticeable drops compared to monolingual baselines.
\section{Language Distribution of Retrieved Documents}
Table~\ref{tab:exposure} reports the language composition of the top-50 documents retrieved for each MKQA query-language split. Across nearly all query languages, the retrieved evidence is heavily concentrated in English, and English often remains the most frequently retrieved language even for non-English queries. This trend is also reflected in the aggregated distribution (\textbf{mkqa\_avg}), indicating that the observed language preference in standard mRAG pipelines largely mirrors structural priors of the retrieval setup rather than purely intrinsic model preference.

\begin{table*}[t]
\centering
\small
\setlength{\tabcolsep}{4.2pt}
\renewcommand{\arraystretch}{1.12}
\resizebox{\textwidth}{!}{%
\begin{tabular}{lllcccccccc}
\toprule
Generator & Level & Method & \textit{en} & \textit{ar} & \textit{es} & zh & \textit{ja} & \textit{de} & \textit{ko} & \textit{th} \\
\midrule

\multirow{6}{*}{Deepseek-chat-v3.1}
& \multirow{4}{*}{Document} & MultiRAG
  & 1.925 & 3.093 & 2.456 & 2.683 & 2.816 & 2.456 & 2.362 & 2.716 \\
&  & CrossRAG
  & 2.005 & 2.897 & 2.706 & 2.822 & 2.968 & 2.328 & 2.778 & 2.653 \\
&  & DKM-RAG
  & 1.955 & 2.710 & 2.262 & 2.572 & 2.841 & 2.227 & 1.733 & 2.843 \\
&  & QTT-RAG
  & \underline{1.671} & 2.347 & 2.054 & \underline{1.348} & 2.475 & \underline{1.663} & \textbf{1.383} & 2.525 \\
& \multirow{2}{*}{Query} & English Translation
  & \multicolumn{1}{c}{--} & \underline{2.170} & \underline{1.992} & 1.800 & \underline{1.857} & 2.224 & 2.179 & \underline{2.110} \\
&  & Ours
  & \textbf{0.851} & \textbf{0.988} & \textbf{1.496} & \textbf{1.288} & \textbf{1.004} & \textbf{0.853} & \underline{1.637} & \textbf{0.955} \\
\midrule

\multirow{6}{*}{Gemini}
& \multirow{4}{*}{Document} & MultiRAG
  & \underline{1.524} & 1.518 & 1.502 & 1.546 & \underline{1.595} & 1.494 & 1.519 & 1.581 \\
&  & CrossRAG
  & 1.688 & 4.821 & \textbf{0.780} & 1.511 & 2.632 & 3.234 & 3.883 & 2.277 \\
&  & DKM-RAG
  & 4.308 & 7.293 & 5.301 & \textbf{0.765} & 6.261 & 5.559 & 6.187 & 9.346 \\
&  & QTT-RAG
  & 2.511 & 7.355 & 5.306 & \underline{1.177} & 6.581 & 5.639 & 6.274 & 9.553 \\
& \multirow{2}{*}{Query} & English Translation
  & \multicolumn{1}{c}{--} & \textbf{1.484} & 1.510 & 1.528 & 1.802 & \underline{1.481} & \textbf{1.478} & \underline{1.539} \\
&  & Ours
  & \textbf{1.483} & \underline{1.485} & \underline{1.469} & 1.484 & \textbf{1.499} & \textbf{1.473} & \underline{1.490} & \textbf{1.481} \\
\midrule

\multirow{6}{*}{Qwen}
& \multirow{4}{*}{Document} & MultiRAG
  & \textbf{1.067} & 1.905 & 1.367 & 1.657 & 1.147 & 1.343 & 1.141 & 1.410 \\
&  & CrossRAG
  & 1.082 & 1.533 & 1.357 & 2.475 & \textbf{1.033} & \textbf{0.630} & \underline{0.693} & 1.543 \\
&  & DKM-RAG
  & 4.573 & 1.368 & 3.794 & 5.171 & 5.721 & \underline{0.764} & \textbf{0.627} & 8.364 \\
&  & QTT-RAG
  & 1.401 & 1.813 & 1.698 & 3.263 & 1.494 & 1.350 & 1.037 & 2.352 \\
& \multirow{2}{*}{Query} & English Translation
  & \multicolumn{1}{c}{--} & \underline{1.086} & \underline{1.347} & \underline{1.141} & 1.095 & 1.108 & 1.214 & \underline{1.226} \\
&  & Ours
  & \underline{1.069} & \textbf{1.036} & \textbf{1.272} & \textbf{1.128} & \underline{1.045} & 1.111 & 1.123 & \textbf{1.225} \\
\bottomrule
\end{tabular}%
}
\caption{Average generation time (sec/query; lower is better). \textbf{Bold} = lowest, \underline{underline} = second-lowest \emph{across both Document-Level and Query-Level rows} for each (generator, language).}
\label{tab:gen_time_full_rows_doc_query_appendix}
\end{table*}

\section{Calibration Details}
Table~\ref{tab:delp_details} reports the detailed DeLP scores for each of the three encoders. After removing the variance explained by the structural priors, the calibrated matrices exhibit a consistent pattern across encoders: the strongest preference concentrates on the diagonal ($L_q{=}L_d$), indicating a robust shift toward query–document language alignment rather than an English-dominant bias. Residual cross-lingual preferences remain comparatively mild and structured, reflecting interpretable related-language effects instead of exposure- or coverage-driven artifacts.
\label{app:delp_details}

\definecolor{heat}{RGB}{255,241,153} 

\providecommand{\best}[1]{\textbf{#1}}
\providecommand{\second}[1]{\underline{#1}} 

\providecommand{\deltasize}{\fontsize{5.4}{5.8}\selectfont}

\providecommand{\rowmin}{0}
\providecommand{\rowmax}{1}
\providecommand{\setrowrange}[2]{\gdef\rowmin{#1}\gdef\rowmax{#2}}

\providecommand{\heatcell}[2]{%
  \begingroup
  \pgfmathsetmacro{\val}{#1}%
  \pgfmathsetmacro{\minval}{\rowmin}%
  \pgfmathsetmacro{\maxval}{\rowmax}%
  \pgfmathsetmacro{\maxint}{60}
  \pgfmathsetmacro{\den}{(\maxval-\minval)}%
  \pgfmathparse{ifthenelse(abs(\den)<0.000001,0,\maxint*(\val-\minval)/\den)}%
  \pgfmathsetmacro{\raw}{\pgfmathresult}%
  \pgfmathsetmacro{\clamped}{max(0,min(\maxint,\raw))}%
  \pgfmathtruncatemacro{\heatpct}{\clamped}%
  \xdef\heatpct{\heatpct}%
  \cellcolor{heat!\heatpct}#2%
  \endgroup
}

\begin{table*}[t]
\centering
\fontsize{7.2}{8.4}\selectfont
\setlength{\tabcolsep}{2.0pt}
\renewcommand{\arraystretch}{1.03}

\begin{tabular}{@{}l l c @{\hspace{2pt}\vrule width 0.4pt\hspace{2pt}} cccccccc@{}}
\toprule
\textbf{Query Lang.} & \textbf{Encoder} &
$\mathbf{L_q=L_d}$ &
\multicolumn{8}{c}{$\mathbf{L_q \neq L_d}$} \\
\cmidrule(lr){4-11}
 &  &  & \textbf{en} & \textbf{ko} & \textbf{zh} & \textbf{fr} & \textbf{ja} & \textbf{it} & \textbf{pt} & \textbf{es} \\
\midrule

\setrowrange{34.58}{49.25}
\multirow{3}{*}{\textbf{en}}
& bge-m3
& \heatcell{49.25}{\best{49.25}}
& \multicolumn{1}{c}{\textcolor{gray}{--}}
& \heatcell{35.87}{35.87 \textcolor{blue}{\deltasize(-13.37)}}
& \heatcell{39.48}{\second{39.48} \textcolor{blue}{\deltasize(-9.77)}}
& \heatcell{34.58}{34.58 \textcolor{blue}{\deltasize(-14.67)}}
& \heatcell{38.79}{38.79 \textcolor{blue}{\deltasize(-10.46)}}
& \heatcell{34.59}{34.59 \textcolor{blue}{\deltasize(-14.66)}}
& \heatcell{35.81}{35.81 \textcolor{blue}{\deltasize(-13.43)}}
& \heatcell{35.13}{35.13 \textcolor{blue}{\deltasize(-14.12)}} \\

\setrowrange{35.83}{50.26}
& p-mMiniLM
& \heatcell{50.26}{\best{50.26}}
& \multicolumn{1}{c}{\textcolor{gray}{--}}
& \heatcell{37.00}{37.00 \textcolor{blue}{\deltasize(-13.27)}}
& \heatcell{40.94}{\second{40.94} \textcolor{blue}{\deltasize(-9.32)}}
& \heatcell{36.18}{36.18 \textcolor{blue}{\deltasize(-14.08)}}
& \heatcell{39.96}{39.96 \textcolor{blue}{\deltasize(-10.31)}}
& \heatcell{35.83}{35.83 \textcolor{blue}{\deltasize(-14.43)}}
& \heatcell{36.62}{36.62 \textcolor{blue}{\deltasize(-13.64)}}
& \heatcell{36.49}{36.49 \textcolor{blue}{\deltasize(-13.78)}} \\

\setrowrange{35.94}{50.80}
& p-mMpNet
& \heatcell{50.80}{\best{50.80}}
& \multicolumn{1}{c}{\textcolor{gray}{--}}
& \heatcell{37.15}{37.15 \textcolor{blue}{\deltasize(-13.65)}}
& \heatcell{40.61}{\second{40.61} \textcolor{blue}{\deltasize(-10.19)}}
& \heatcell{35.94}{35.94 \textcolor{blue}{\deltasize(-14.86)}}
& \heatcell{40.09}{40.09 \textcolor{blue}{\deltasize(-10.71)}}
& \heatcell{36.04}{36.04 \textcolor{blue}{\deltasize(-14.76)}}
& \heatcell{36.96}{36.96 \textcolor{blue}{\deltasize(-13.84)}}
& \heatcell{36.43}{36.43 \textcolor{blue}{\deltasize(-14.37)}} \\

\midrule

\setrowrange{34.41}{42.45}
\multirow{3}{*}{\textbf{ko}}
& bge-m3
& \heatcell{42.34}{\second{42.34}}
& \heatcell{36.76}{36.76 \textcolor{blue}{\deltasize(-5.58)}}
& \multicolumn{1}{c}{\textcolor{gray}{--}}
& \heatcell{40.69}{40.69 \textcolor{blue}{\deltasize(-1.65)}}
& \heatcell{34.41}{34.41 \textcolor{blue}{\deltasize(-7.93)}}
& \heatcell{42.45}{\best{42.45} \textcolor{red}{\deltasize(+0.11)}}
& \heatcell{34.44}{34.44 \textcolor{blue}{\deltasize(-7.90)}}
& \heatcell{35.29}{35.29 \textcolor{blue}{\deltasize(-7.05)}}
& \heatcell{34.46}{34.46 \textcolor{blue}{\deltasize(-7.87)}} \\

\setrowrange{34.75}{44.09}
& p-mMiniLM
& \heatcell{44.09}{\best{44.09}}
& \heatcell{38.03}{38.03 \textcolor{blue}{\deltasize(-6.06)}}
& \multicolumn{1}{c}{\textcolor{gray}{--}}
& \heatcell{42.37}{\second{42.37} \textcolor{blue}{\deltasize(-1.72)}}
& \heatcell{35.09}{35.09 \textcolor{blue}{\deltasize(-9.00)}}
& \heatcell{43.90}{43.90 \textcolor{blue}{\deltasize(-0.19)}}
& \heatcell{34.75}{34.75 \textcolor{blue}{\deltasize(-9.34)}}
& \heatcell{36.07}{36.07 \textcolor{blue}{\deltasize(-8.02)}}
& \heatcell{34.98}{34.98 \textcolor{blue}{\deltasize(-9.11)}} \\

\setrowrange{34.87}{44.43}
& p-mMpNet
& \heatcell{43.72}{\second{43.72}}
& \heatcell{38.29}{38.29 \textcolor{blue}{\deltasize(-5.43)}}
& \multicolumn{1}{c}{\textcolor{gray}{--}}
& \heatcell{42.19}{42.19 \textcolor{blue}{\deltasize(-1.53)}}
& \heatcell{35.20}{35.20 \textcolor{blue}{\deltasize(-8.52)}}
& \heatcell{44.43}{\best{44.43} \textcolor{red}{\deltasize(+0.72)}}
& \heatcell{34.91}{34.91 \textcolor{blue}{\deltasize(-8.80)}}
& \heatcell{35.59}{35.59 \textcolor{blue}{\deltasize(-8.13)}}
& \heatcell{34.87}{34.87 \textcolor{blue}{\deltasize(-8.84)}} \\

\midrule

\setrowrange{34.33}{49.67}
\multirow{3}{*}{\textbf{zh}}
& bge-m3
& \heatcell{49.67}{\best{49.67}}
& \heatcell{38.52}{38.52 \textcolor{blue}{\deltasize(-11.15)}}
& \heatcell{37.32}{37.32 \textcolor{blue}{\deltasize(-12.36)}}
& \multicolumn{1}{c}{\textcolor{gray}{--}}
& \heatcell{34.33}{34.33 \textcolor{blue}{\deltasize(-15.34)}}
& \heatcell{41.36}{\second{41.36} \textcolor{blue}{\deltasize(-8.32)}}
& \heatcell{34.58}{34.58 \textcolor{blue}{\deltasize(-15.09)}}
& \heatcell{35.71}{35.71 \textcolor{blue}{\deltasize(-13.96)}}
& \heatcell{34.98}{34.98 \textcolor{blue}{\deltasize(-14.69)}} \\

\setrowrange{34.99}{51.01}
& p-mMiniLM
& \heatcell{51.01}{\best{51.01}}
& \heatcell{38.80}{38.80 \textcolor{blue}{\deltasize(-12.21)}}
& \heatcell{38.11}{38.11 \textcolor{blue}{\deltasize(-12.90)}}
& \multicolumn{1}{c}{\textcolor{gray}{--}}
& \heatcell{34.99}{34.99 \textcolor{blue}{\deltasize(-16.03)}}
& \heatcell{42.20}{\second{42.20} \textcolor{blue}{\deltasize(-8.81)}}
& \heatcell{35.06}{35.06 \textcolor{blue}{\deltasize(-15.95)}}
& \heatcell{35.94}{35.94 \textcolor{blue}{\deltasize(-15.07)}}
& \heatcell{35.38}{35.38 \textcolor{blue}{\deltasize(-15.64)}} \\

\setrowrange{34.87}{51.11}
& p-mMpNet
& \heatcell{51.11}{\best{51.11}}
& \heatcell{38.72}{38.72 \textcolor{blue}{\deltasize(-12.39)}}
& \heatcell{37.91}{37.91 \textcolor{blue}{\deltasize(-13.20)}}
& \multicolumn{1}{c}{\textcolor{gray}{--}}
& \heatcell{34.87}{34.87 \textcolor{blue}{\deltasize(-16.24)}}
& \heatcell{42.13}{\second{42.13} \textcolor{blue}{\deltasize(-8.98)}}
& \heatcell{34.98}{34.98 \textcolor{blue}{\deltasize(-16.13)}}
& \heatcell{35.88}{35.88 \textcolor{blue}{\deltasize(-15.23)}}
& \heatcell{35.31}{35.31 \textcolor{blue}{\deltasize(-15.80)}} \\

\midrule

\setrowrange{35.39}{40.48}
\multirow{3}{*}{\textbf{fr}}
& bge-m3
& \heatcell{39.45}{39.45}
& \heatcell{40.48}{\best{40.48} \textcolor{red}{\deltasize(+1.02)}}
& \heatcell{36.15}{36.15 \textcolor{blue}{\deltasize(-3.30)}}
& \heatcell{39.95}{\second{39.95} \textcolor{red}{\deltasize(+0.49)}}
& \multicolumn{1}{c}{\textcolor{gray}{--}}
& \heatcell{39.47}{39.47 \textcolor{red}{\deltasize(+0.01)}}
& \heatcell{35.39}{35.39 \textcolor{blue}{\deltasize(-4.06)}}
& \heatcell{36.25}{36.25 \textcolor{blue}{\deltasize(-3.20)}}
& \heatcell{35.75}{35.75 \textcolor{blue}{\deltasize(-3.70)}} \\

\setrowrange{36.33}{41.56}
& p-mMiniLM
& \heatcell{40.42}{40.42}
& \heatcell{41.56}{\best{41.56} \textcolor{red}{\deltasize(+1.14)}}
& \heatcell{37.20}{37.20 \textcolor{blue}{\deltasize(-3.22)}}
& \heatcell{40.85}{\second{40.85} \textcolor{red}{\deltasize(+0.43)}}
& \multicolumn{1}{c}{\textcolor{gray}{--}}
& \heatcell{40.27}{40.27 \textcolor{blue}{\deltasize(-0.15)}}
& \heatcell{36.33}{36.33 \textcolor{blue}{\deltasize(-4.09)}}
& \heatcell{36.94}{36.94 \textcolor{blue}{\deltasize(-3.48)}}
& \heatcell{36.56}{36.56 \textcolor{blue}{\deltasize(-3.86)}} \\

\setrowrange{36.29}{41.45}
& p-mMpNet
& \heatcell{40.28}{40.28}
& \heatcell{41.45}{\best{41.45} \textcolor{red}{\deltasize(+1.17)}}
& \heatcell{36.95}{36.95 \textcolor{blue}{\deltasize(-3.33)}}
& \heatcell{40.71}{\second{40.71} \textcolor{red}{\deltasize(+0.43)}}
& \multicolumn{1}{c}{\textcolor{gray}{--}}
& \heatcell{40.03}{40.03 \textcolor{blue}{\deltasize(-0.25)}}
& \heatcell{36.29}{36.29 \textcolor{blue}{\deltasize(-3.98)}}
& \heatcell{36.87}{36.87 \textcolor{blue}{\deltasize(-3.41)}}
& \heatcell{36.54}{36.54 \textcolor{blue}{\deltasize(-3.74)}} \\

\midrule

\setrowrange{34.70}{48.61}
\multirow{3}{*}{\textbf{ja}}
& bge-m3
& \heatcell{48.61}{\best{48.61}}
& \heatcell{38.44}{38.44 \textcolor{blue}{\deltasize(-10.16)}}
& \heatcell{38.21}{38.21 \textcolor{blue}{\deltasize(-10.39)}}
& \heatcell{41.15}{\second{41.15} \textcolor{blue}{\deltasize(-7.46)}}
& \heatcell{34.70}{34.70 \textcolor{blue}{\deltasize(-13.91)}}
& \multicolumn{1}{c}{\textcolor{gray}{--}}
& \heatcell{34.83}{34.83 \textcolor{blue}{\deltasize(-13.78)}}
& \heatcell{35.86}{35.86 \textcolor{blue}{\deltasize(-12.75)}}
& \heatcell{35.09}{35.09 \textcolor{blue}{\deltasize(-13.52)}} \\

\setrowrange{35.19}{49.56}
& p-mMiniLM
& \heatcell{49.56}{\best{49.56}}
& \heatcell{38.95}{38.95 \textcolor{blue}{\deltasize(-10.61)}}
& \heatcell{38.55}{38.55 \textcolor{blue}{\deltasize(-11.01)}}
& \heatcell{41.90}{\second{41.90} \textcolor{blue}{\deltasize(-7.66)}}
& \heatcell{35.19}{35.19 \textcolor{blue}{\deltasize(-14.37)}}
& \multicolumn{1}{c}{\textcolor{gray}{--}}
& \heatcell{35.21}{35.21 \textcolor{blue}{\deltasize(-14.35)}}
& \heatcell{36.14}{36.14 \textcolor{blue}{\deltasize(-13.42)}}
& \heatcell{35.44}{35.44 \textcolor{blue}{\deltasize(-14.12)}} \\

\setrowrange{34.94}{49.41}
& p-mMpNet
& \heatcell{49.41}{\best{49.41}}
& \heatcell{38.70}{38.70 \textcolor{blue}{\deltasize(-10.72)}}
& \heatcell{38.43}{38.43 \textcolor{blue}{\deltasize(-10.98)}}
& \heatcell{41.64}{\second{41.64} \textcolor{blue}{\deltasize(-7.78)}}
& \heatcell{34.94}{34.94 \textcolor{blue}{\deltasize(-14.47)}}
& \multicolumn{1}{c}{\textcolor{gray}{--}}
& \heatcell{34.94}{34.94 \textcolor{blue}{\deltasize(-14.47)}}
& \heatcell{35.92}{35.92 \textcolor{blue}{\deltasize(-13.49)}}
& \heatcell{35.15}{35.15 \textcolor{blue}{\deltasize(-14.26)}} \\

\midrule

\setrowrange{35.87}{39.88}
\multirow{3}{*}{\textbf{it}}
& bge-m3
& \heatcell{38.38}{38.38}
& \heatcell{39.88}{\best{39.88} \textcolor{red}{\deltasize(+1.50)}}
& \heatcell{36.15}{36.15 \textcolor{blue}{\deltasize(-2.23)}}
& \heatcell{39.83}{\second{39.83} \textcolor{red}{\deltasize(+1.45)}}
& \heatcell{35.87}{35.87 \textcolor{blue}{\deltasize(-2.51)}}
& \heatcell{39.26}{39.26 \textcolor{red}{\deltasize(+0.88)}}
& \multicolumn{1}{c}{\textcolor{gray}{--}}
& \heatcell{36.39}{36.39 \textcolor{blue}{\deltasize(-2.00)}}
& \heatcell{36.17}{36.17 \textcolor{blue}{\deltasize(-2.21)}} \\

\setrowrange{37.08}{41.10}
& p-mMiniLM
& \heatcell{39.44}{39.44}
& \heatcell{41.10}{\best{41.10} \textcolor{red}{\deltasize(+1.67)}}
& \heatcell{37.23}{37.23 \textcolor{blue}{\deltasize(-2.21)}}
& \heatcell{40.92}{\second{40.92} \textcolor{red}{\deltasize(+1.48)}}
& \heatcell{37.08}{37.08 \textcolor{blue}{\deltasize(-2.36)}}
& \heatcell{40.24}{40.24 \textcolor{red}{\deltasize(+0.80)}}
& \multicolumn{1}{c}{\textcolor{gray}{--}}
& \heatcell{37.44}{37.44 \textcolor{blue}{\deltasize(-2.00)}}
& \heatcell{37.36}{37.36 \textcolor{blue}{\deltasize(-2.08)}} \\

\setrowrange{36.94}{41.02}
& p-mMpNet
& \heatcell{39.33}{39.33}
& \heatcell{40.90}{\second{40.90} \textcolor{red}{\deltasize(+1.57)}}
& \heatcell{37.18}{37.18 \textcolor{blue}{\deltasize(-2.15)}}
& \heatcell{41.02}{\best{41.02} \textcolor{red}{\deltasize(+1.69)}}
& \heatcell{36.94}{36.94 \textcolor{blue}{\deltasize(-2.39)}}
& \heatcell{40.09}{40.09 \textcolor{red}{\deltasize(+0.77)}}
& \multicolumn{1}{c}{\textcolor{gray}{--}}
& \heatcell{37.21}{37.21 \textcolor{blue}{\deltasize(-2.12)}}
& \heatcell{37.20}{37.20 \textcolor{blue}{\deltasize(-2.12)}} \\

\midrule

\setrowrange{35.78}{45.38}
\multirow{3}{*}{\textbf{pt}}
& bge-m3
& \heatcell{45.38}{\best{45.38}}
& \heatcell{39.89}{\second{39.89} \textcolor{blue}{\deltasize(-5.49)}}
& \heatcell{36.22}{36.22 \textcolor{blue}{\deltasize(-9.17)}}
& \heatcell{39.82}{39.82 \textcolor{blue}{\deltasize(-5.56)}}
& \heatcell{35.78}{35.78 \textcolor{blue}{\deltasize(-9.60)}}
& \heatcell{39.41}{39.41 \textcolor{blue}{\deltasize(-5.97)}}
& \heatcell{35.81}{35.81 \textcolor{blue}{\deltasize(-9.57)}}
& \multicolumn{1}{c}{\textcolor{gray}{--}}
& \heatcell{37.09}{37.09 \textcolor{blue}{\deltasize(-8.29)}} \\

\setrowrange{36.93}{46.52}
& p-mMiniLM
& \heatcell{46.52}{\best{46.52}}
& \heatcell{41.16}{41.16 \textcolor{blue}{\deltasize(-5.36)}}
& \heatcell{37.33}{37.33 \textcolor{blue}{\deltasize(-9.19)}}
& \heatcell{41.24}{\second{41.24} \textcolor{blue}{\deltasize(-5.28)}}
& \heatcell{37.03}{37.03 \textcolor{blue}{\deltasize(-9.49)}}
& \heatcell{40.47}{40.47 \textcolor{blue}{\deltasize(-6.05)}}
& \heatcell{36.93}{36.93 \textcolor{blue}{\deltasize(-9.59)}}
& \multicolumn{1}{c}{\textcolor{gray}{--}}
& \heatcell{38.21}{38.21 \textcolor{blue}{\deltasize(-8.31)}} \\

\setrowrange{36.70}{46.33}
& p-mMpNet
& \heatcell{46.33}{\best{46.33}}
& \heatcell{40.61}{\second{40.61} \textcolor{blue}{\deltasize(-5.72)}}
& \heatcell{37.38}{37.38 \textcolor{blue}{\deltasize(-8.95)}}
& \heatcell{40.84}{40.84 \textcolor{blue}{\deltasize(-5.49)}}
& \heatcell{36.70}{36.70 \textcolor{blue}{\deltasize(-9.63)}}
& \heatcell{40.14}{40.14 \textcolor{blue}{\deltasize(-6.19)}}
& \heatcell{36.71}{36.71 \textcolor{blue}{\deltasize(-9.61)}}
& \multicolumn{1}{c}{\textcolor{gray}{--}}
& \heatcell{37.88}{37.88 \textcolor{blue}{\deltasize(-8.45)}} \\

\midrule

\setrowrange{35.68}{40.18}
\multirow{3}{*}{\textbf{es}}
& bge-m3
& \heatcell{37.64}{37.64}
& \heatcell{40.18}{\best{40.18} \textcolor{red}{\deltasize(+2.54)}}
& \heatcell{36.21}{36.21 \textcolor{blue}{\deltasize(-1.43)}}
& \heatcell{39.78}{\second{39.78} \textcolor{red}{\deltasize(+2.15)}}
& \heatcell{35.68}{35.68 \textcolor{blue}{\deltasize(-1.95)}}
& \heatcell{39.28}{39.28 \textcolor{red}{\deltasize(+1.64)}}
& \heatcell{35.90}{35.90 \textcolor{blue}{\deltasize(-1.74)}}
& \heatcell{36.82}{36.82 \textcolor{blue}{\deltasize(-0.82)}}
& \multicolumn{1}{c}{\textcolor{gray}{--}} \\

\setrowrange{36.87}{41.31}
& p-mMiniLM
& \heatcell{38.71}{38.71}
& \heatcell{41.31}{\best{41.31} \textcolor{red}{\deltasize(+2.60)}}
& \heatcell{37.29}{37.29 \textcolor{blue}{\deltasize(-1.42)}}
& \heatcell{40.85}{\second{40.85} \textcolor{red}{\deltasize(+2.14)}}
& \heatcell{36.87}{36.87 \textcolor{blue}{\deltasize(-1.84)}}
& \heatcell{40.20}{40.20 \textcolor{red}{\deltasize(+1.49)}}
& \heatcell{37.01}{37.01 \textcolor{blue}{\deltasize(-1.70)}}
& \heatcell{37.73}{37.73 \textcolor{blue}{\deltasize(-0.98)}}
& \multicolumn{1}{c}{\textcolor{gray}{--}} \\

\setrowrange{36.34}{40.65}
& p-mMpNet
& \heatcell{38.23}{38.23}
& \heatcell{40.65}{\best{40.65} \textcolor{red}{\deltasize(+2.42)}}
& \heatcell{37.09}{37.09 \textcolor{blue}{\deltasize(-1.14)}}
& \heatcell{40.53}{\second{40.53} \textcolor{red}{\deltasize(+2.30)}}
& \heatcell{36.34}{36.34 \textcolor{blue}{\deltasize(-1.89)}}
& \heatcell{39.81}{39.81 \textcolor{red}{\deltasize(+1.58)}}
& \heatcell{36.43}{36.43 \textcolor{blue}{\deltasize(-1.80)}}
& \heatcell{37.19}{37.19 \textcolor{blue}{\deltasize(-1.04)}}
& \multicolumn{1}{c}{\textcolor{gray}{--}} \\

\bottomrule
\end{tabular}
\caption{Language preference measured by DeLP. Each cell reports the debiased preference score and its delta from the matching-language baseline ($L_q=L_d$). Background shading is row-wise min--max scaled (including the diagonal cell); Darker cells indicate a stronger preference for the document language.}

\vspace{-5mm}
\label{tab:delp_details}
\end{table*}

\section{Gold Passage Counting Protocol}
\label{app:gold}

We compute Table~\ref{tab:rq1_single_table_nodelta} on the 2,827-question subset that overlaps with KILT NQ provenance. Because prior work~\cite{park2025investigating} provides the same underlying question translated into 13 query languages, the unit counted in Table~\ref{tab:rq1_single_table_nodelta} is not the number of unique questions but the number of question$\times$query-language instances. Hence the total number of instances is $2{,}827\times 13=36{,}751$, and values such as ``\#q = 26{,}934 (73.29\%)'' can legitimately exceed 2,827; the ratio is computed as $26{,}934/36{,}751$. Gold labels originate from KILT’s provenance, which is anchored to English Wikipedia page IDs (WPIDs). All 13 translations of the same question share the same gold WPID set. We then assess gold availability in each Wikipedia language edition by mapping each English WPID to a corresponding page in language $\ell$ using Wikipedia/Wikidata interlanguage links (sitelinks), and checking whether the mapped page exists in the Wikipedia dump used to build our corpus. 

A key source of confusion is that our corpus is passage-based: each Wikipedia page is split into multiple chunks, so a single WPID may correspond to multiple passages. Moreover, for a given question, KILT may provide multiple gold provenances that map to different passages within the same WPID. In Table~\ref{tab:rq1_single_table_nodelta}, we use a WPID-level convention: when multiple gold passages correspond to the same WPID for a query, we treat them as a single gold item rather than counting them multiple times. Out of the 2,827 questions in our KILT-overlap subset, 2,404 questions have at least one available gold WPID (i.e., a mapped gold page exists in our Wikipedia dumps). Accordingly, the number of questions with gold evidence satisfies \texttt{only\_en} + \texttt{both} = 2{,}404 in Table~\ref{tab:rq1_single_table_nodelta}.

\section{Detailed Latency}
\label{app:latency}
Table~\ref{tab:gen_time_full_rows_doc_query_appendix} reports detailed latency measured as average generation time (sec/query; lower is better) for each generator across languages.
DELTA remains consistently efficient because it produces a single fused query and avoids document translation overhead; in several settings, it is even faster than English Translation, since the fused query retains local cues and disambiguation anchors that help retrieval focus earlier and reduce wasted computation on overly generic English-only signals.

\section{Cultural Prior Measurement}
\label{app:cultural}
To model whether a query is intrinsically tied to a particular cultural or regional context (independent of corpus size or retrieval exposure), we construct a \emph{cultural prior} using an LLM-based classifier. Starting from the English version of each query (MKQA-en), we assign exactly one \emph{cultural database language} from a fixed set of 13 languages (\textit{en, ar, es, de, ja, ko, th, zh, fr, it, pt, ru, fi}). We use GPT-4o mini (via OpenRouter) with constrained JSON output to enforce a single-label decision. We instruct the classifier to choose the local language of the primary place/culture the query is about (e.g., France $\rightarrow$ \textit{fr}, Hong Kong/China $\rightarrow$ \textit{zh}), and to select \textit{en} only when the cultural context is inherently English-speaking (e.g., US/UK-specific) or when the query is genuinely global/multi-country and not tied to a single locale.

In addition to the cultural-language label, we record lightweight \emph{cultural metadata} for analysis and filtering:
(i) \texttt{country\_or\_region} (a single primary place/region),
(ii) \texttt{is\_culture\_specific} (whether the question is judged to be culture/locale-specific),
(iii) \texttt{confidence} (0--1), and
(iv) a short \texttt{rationale}.
These fields are used only to characterize the dataset and to support qualitative inspection; our core metric relies on the language label.

Finally, we define the cultural prior $p_{\text{cult}}(\ell)$ as the empirical probability that a query’s predicted cultural language equals $\ell$, i.e., the normalized frequency of the single-label assignments over the evaluation set. This prior captures \emph{where evidence should exist} in a fair localized setting, and is incorporated as a structural factor alongside other priors (e.g., exposure and gold availability) in our calibration analysis. We use those cultural prior and metadata for calibration and DELTA.

\section{Repetition-based weighting for DELTA}
\label{app:delta_repetition}

\paragraph{Query construction with repetition.}
To control cue influence without changing the retriever or learning parameters, we apply a deterministic repetition policy while constructing the fused query string $Q_{\text{fused}}$.
We use the same notation as in Eq.~\ref{eq:delta_repetition_glob}: $y\in\{0,1\}$ indicates whether the query is culture-specific and $c\in[0,1]$ is the confidence score.
We repeat the local block \texttt{[LOCAL:$L_q$]} $r_{\text{local}}$ times and the global pivot block \texttt{[GLOB]} $r_{\text{glob}}$ times, and concatenate all segments with a delimiter (``\texttt{ | }'') to form a single retrieval query.

\paragraph{Length control.}
To keep retrieval budgets comparable across methods, we truncate the final $Q_{\text{fused}}$ to a fixed maximum length (e.g., 900 characters) after concatenation.

\paragraph{Deduplication.}
We apply conservative deduplication to avoid redundant anchors:
(i) if the global and local titles are identical, we keep only a single \texttt{[TITLE\_BRIDGE]};
(ii) if alias sets match across languages, we keep only the global alias block;
and (iii) when the query language is not English, we always include \texttt{[LOCAL:$L_q$]} at least once.

\begin{table*}[!ht]
\centering
\setlength{\tabcolsep}{3pt}
\renewcommand{\arraystretch}{1.0}
\resizebox{0.95\textwidth}{!}{
\begin{tabular}{cc|>{\columncolor{gray!15}}c|cccccccc}
\toprule
\multicolumn{2}{c|}{} 
& \textbf{\(L_q = L_d\)} 
& \multicolumn{8}{c}{\textbf{\(L_q \neq L_d\)}} \\
\textbf{Query Lang.} & \textbf{Encoder}
& 
& \textbf{en} & \textbf{ko} & \textbf{zh} & \textbf{fr} & \textbf{ja} & \textbf{it} & \textbf{pt} & \textbf{es} \\
\midrule
\multirow{3}{*}{\textbf{en}}
& \textbf{bge-m3}
  & \textbf{56.03} 
  & -- 
  & 33.02 {\scriptsize(\textcolor{blue}{-23.01})}
  & 33.10 {\scriptsize(\textcolor{blue}{-22.93})}
  & 36.61 {\scriptsize(\textcolor{blue}{-19.42})}
  & 33.36 {\scriptsize(\textcolor{blue}{-22.67})}
  & 35.89 {\scriptsize(\textcolor{blue}{-20.14})}
  & 35.86 {\scriptsize(\textcolor{blue}{-20.17})}
  & \underline{36.62} {\scriptsize(\textcolor{blue}{-19.41})} \\
& \textbf{p-mMiniLM}
  & \textbf{56.85}
  & -- 
  & 34.34 {\scriptsize(\textcolor{blue}{-22.51})}
  & 34.61 {\scriptsize(\textcolor{blue}{-22.24})}
  & \underline{38.17} {\scriptsize(\textcolor{blue}{-18.68})}
  & 34.52 {\scriptsize(\textcolor{blue}{-22.33})}
  & 37.15 {\scriptsize(\textcolor{blue}{-19.70})}
  & 36.73 {\scriptsize(\textcolor{blue}{-20.12})}
  & 37.96 {\scriptsize(\textcolor{blue}{-18.89})} \\
& \textbf{p-mMpNet}
  & \textbf{57.49}
  & -- 
  & 34.45 {\scriptsize(\textcolor{blue}{-23.04})}
  & 34.27 {\scriptsize(\textcolor{blue}{-23.22})}
  & \underline{37.94} {\scriptsize(\textcolor{blue}{-19.55})}
  & 34.67 {\scriptsize(\textcolor{blue}{-22.82})}
  & 37.34 {\scriptsize(\textcolor{blue}{-20.15})}
  & 37.02 {\scriptsize(\textcolor{blue}{-20.47})}
  & 37.90 {\scriptsize(\textcolor{blue}{-19.59})} \\
\midrule
\multirow{3}{*}{\textbf{ko}}
& \textbf{bge-m3}
  & \underline{41.15} 
  & \textbf{43.49} {\scriptsize(\textcolor{red}{+2.34})}
  & -- 
  & 34.42 {\scriptsize(\textcolor{blue}{-6.73})}
  & 36.42 {\scriptsize(\textcolor{blue}{-4.73})}
  & 37.18 {\scriptsize(\textcolor{blue}{-3.97})}
  & 35.72 {\scriptsize(\textcolor{blue}{-5.43})}
  & 35.30 {\scriptsize(\textcolor{blue}{-5.85})}
  & 35.93 {\scriptsize(\textcolor{blue}{-5.22})} \\
& \textbf{p-mMiniLM}
  & \underline{42.95}
  & \textbf{44.62} {\scriptsize(\textcolor{red}{+1.67})}
  & -- 
  & 36.04 {\scriptsize(\textcolor{blue}{-6.91})}
  & 37.08 {\scriptsize(\textcolor{blue}{-5.87})}
  & 38.47 {\scriptsize(\textcolor{blue}{-4.48})}
  & 36.07 {\scriptsize(\textcolor{blue}{-6.88})}
  & 36.18 {\scriptsize(\textcolor{blue}{-6.77})}
  & 36.45 {\scriptsize(\textcolor{blue}{-6.50})} \\
& \textbf{p-mMpNet}
  & \underline{42.53}
  & \textbf{44.98} {\scriptsize(\textcolor{red}{+2.45})}
  & -- 
  & 35.85 {\scriptsize(\textcolor{blue}{-6.68})}
  & 37.20 {\scriptsize(\textcolor{blue}{-5.33})}
  & 39.01 {\scriptsize(\textcolor{blue}{-3.52})}
  & 36.21 {\scriptsize(\textcolor{blue}{-6.32})}
  & 35.65 {\scriptsize(\textcolor{blue}{-6.88})}
  & 36.34 {\scriptsize(\textcolor{blue}{-6.19})} \\
\midrule
\multirow{3}{*}{\textbf{zh}}
& \textbf{bge-m3}
  & \underline{44.98}
  & \textbf{45.26} {\scriptsize(\textcolor{red}{+0.28})}
  & 34.52 {\scriptsize(\textcolor{blue}{-10.46})}
  & -- 
  & 36.34 {\scriptsize(\textcolor{blue}{-8.64})}
  & 36.05 {\scriptsize(\textcolor{blue}{-8.93})}
  & 35.86 {\scriptsize(\textcolor{blue}{-9.12})}
  & 35.73 {\scriptsize(\textcolor{blue}{-9.25})}
  & 36.45 {\scriptsize(\textcolor{blue}{-8.53})} \\
& \textbf{p-mMiniLM}
  & \textbf{46.18}
  & \underline{45.39} {\scriptsize(\textcolor{blue}{-0.79})}
  & 35.46 {\scriptsize(\textcolor{blue}{-10.72})}
  & -- 
  & 36.98 {\scriptsize(\textcolor{blue}{-9.20})}
  & 36.77 {\scriptsize(\textcolor{blue}{-9.41})}
  & 36.38 {\scriptsize(\textcolor{blue}{-9.80})}
  & 36.05 {\scriptsize(\textcolor{blue}{-10.13})}
  & 36.85 {\scriptsize(\textcolor{blue}{-9.33})} \\
& \textbf{p-mMpNet}
  & \textbf{46.27}
  & \underline{45.41} {\scriptsize(\textcolor{blue}{-0.86})}
  & 35.21 {\scriptsize(\textcolor{blue}{-11.06})}
  & -- 
  & 36.87 {\scriptsize(\textcolor{blue}{-9.40})}
  & 36.71 {\scriptsize(\textcolor{blue}{-9.56})}
  & 36.28 {\scriptsize(\textcolor{blue}{-9.99})}
  & 35.94 {\scriptsize(\textcolor{blue}{-10.33})}
  & 36.78 {\scriptsize(\textcolor{blue}{-9.49})} \\
\midrule
\multirow{3}{*}{\textbf{fr}}
& \textbf{bge-m3}
  & \underline{43.18}
  & \textbf{47.23} {\scriptsize(\textcolor{red}{+4.05})}
  & 33.29 {\scriptsize(\textcolor{blue}{-9.89})}
  & 33.58 {\scriptsize(\textcolor{blue}{-9.60})}
  & -- 
  & 34.07 {\scriptsize(\textcolor{blue}{-9.11})}
  & 36.70 {\scriptsize(\textcolor{blue}{-6.48})}
  & 36.30 {\scriptsize(\textcolor{blue}{-6.88})}
  & 37.25 {\scriptsize(\textcolor{blue}{-5.93})} \\
& \textbf{p-mMiniLM}
  & \underline{44.09}
  & \textbf{48.15} {\scriptsize(\textcolor{red}{+4.06})}
  & 34.54 {\scriptsize(\textcolor{blue}{-9.55})}
  & 34.52 {\scriptsize(\textcolor{blue}{-9.57})}
  & -- 
  & 34.83 {\scriptsize(\textcolor{blue}{-9.26})}
  & 37.65 {\scriptsize(\textcolor{blue}{-6.44})}
  & 37.05 {\scriptsize(\textcolor{blue}{-7.04})}
  & 38.03 {\scriptsize(\textcolor{blue}{-6.06})} \\
& \textbf{p-mMpNet}
  & \underline{43.96}
  & \textbf{48.14} {\scriptsize(\textcolor{red}{+4.18})}
  & 34.25 {\scriptsize(\textcolor{blue}{-9.71})}
  & 34.37 {\scriptsize(\textcolor{blue}{-9.59})}
  & -- 
  & 34.61 {\scriptsize(\textcolor{blue}{-9.35})}
  & 37.59 {\scriptsize(\textcolor{blue}{-6.37})}
  & 36.93 {\scriptsize(\textcolor{blue}{-7.03})}
  & 38.01 {\scriptsize(\textcolor{blue}{-5.95})} \\
\midrule
\multirow{3}{*}{\textbf{ja}}
& \textbf{bge-m3}
  & \underline{45.03}
  & \textbf{45.18} {\scriptsize(\textcolor{red}{+0.15})}
  & 35.45 {\scriptsize(\textcolor{blue}{-9.58})}
  & 34.86 {\scriptsize(\textcolor{blue}{-10.17})}
  & 36.71 {\scriptsize(\textcolor{blue}{-8.32})}
  & -- 
  & 36.11 {\scriptsize(\textcolor{blue}{-8.92})}
  & 35.88 {\scriptsize(\textcolor{blue}{-9.15})}
  & 36.56 {\scriptsize(\textcolor{blue}{-8.47})} \\
& \textbf{p-mMiniLM}
  & \textbf{45.80}
  & \underline{45.54} {\scriptsize(\textcolor{blue}{-0.26})}
  & 35.90 {\scriptsize(\textcolor{blue}{-9.90})}
  & 35.57 {\scriptsize(\textcolor{blue}{-10.23})}
  & 37.18 {\scriptsize(\textcolor{blue}{-8.62})}
  & -- 
  & 36.53 {\scriptsize(\textcolor{blue}{-9.27})}
  & 36.25 {\scriptsize(\textcolor{blue}{-9.55})}
  & 36.91 {\scriptsize(\textcolor{blue}{-8.89})} \\
& \textbf{p-mMpNet}
  & \textbf{45.67}
  & \underline{45.39} {\scriptsize(\textcolor{blue}{-0.28})}
  & 35.73 {\scriptsize(\textcolor{blue}{-9.94})}
  & 35.30 {\scriptsize(\textcolor{blue}{-10.37})}
  & 36.94 {\scriptsize(\textcolor{blue}{-8.73})}
  & -- 
  & 36.24 {\scriptsize(\textcolor{blue}{-9.43})}
  & 35.98 {\scriptsize(\textcolor{blue}{-9.69})}
  & 36.62 {\scriptsize(\textcolor{blue}{-9.05})} \\
\midrule
\multirow{3}{*}{\textbf{it}}
& \textbf{bge-m3}
  & \underline{41.06}
  & \textbf{46.63} {\scriptsize(\textcolor{red}{+5.57})}
  & 33.30 {\scriptsize(\textcolor{blue}{-7.76})}
  & 33.47 {\scriptsize(\textcolor{blue}{-7.59})}
  & 37.92 {\scriptsize(\textcolor{blue}{-3.14})}
  & 33.86 {\scriptsize(\textcolor{blue}{-7.20})}
  & -- 
  & 36.44 {\scriptsize(\textcolor{blue}{-4.62})}
  & 37.68 {\scriptsize(\textcolor{blue}{-3.38})} \\
& \textbf{p-mMiniLM}
  & \underline{42.11}
  & \textbf{47.69} {\scriptsize(\textcolor{red}{+5.58})}
  & 34.57 {\scriptsize(\textcolor{blue}{-7.54})}
  & 34.59 {\scriptsize(\textcolor{blue}{-7.52})}
  & 39.07 {\scriptsize(\textcolor{blue}{-3.04})}
  & 34.80 {\scriptsize(\textcolor{blue}{-7.31})}
  & -- 
  & 37.55 {\scriptsize(\textcolor{blue}{-4.56})}
  & 38.83 {\scriptsize(\textcolor{blue}{-3.28})} \\
& \textbf{p-mMpNet}
  & \underline{41.98}
  & \textbf{47.59} {\scriptsize(\textcolor{red}{+5.61})}
  & 34.48 {\scriptsize(\textcolor{blue}{-7.50})}
  & 34.68 {\scriptsize(\textcolor{blue}{-7.30})}
  & 38.94 {\scriptsize(\textcolor{blue}{-3.04})}
  & 34.67 {\scriptsize(\textcolor{blue}{-7.31})}
  & -- 
  & 37.27 {\scriptsize(\textcolor{blue}{-4.71})}
  & 38.67 {\scriptsize(\textcolor{blue}{-3.31})} \\
\midrule
\multirow{3}{*}{\textbf{pt}}
& \textbf{bge-m3}
  & \underline{39.19}
  & \textbf{46.64} {\scriptsize(\textcolor{red}{+7.45})}
  & 33.37 {\scriptsize(\textcolor{blue}{-5.82})}
  & 33.46 {\scriptsize(\textcolor{blue}{-5.73})}
  & 37.83 {\scriptsize(\textcolor{blue}{-1.36})}
  & 34.02 {\scriptsize(\textcolor{blue}{-5.17})}
  & 37.13 {\scriptsize(\textcolor{blue}{-2.06})}
  & -- 
  & 38.61 {\scriptsize(\textcolor{blue}{-0.58})} \\
& \textbf{p-mMiniLM}
  & \underline{40.17}
  & \textbf{47.75} {\scriptsize(\textcolor{red}{+7.58})}
  & 34.67 {\scriptsize(\textcolor{blue}{-5.50})}
  & 34.91 {\scriptsize(\textcolor{blue}{-5.26})}
  & 39.02 {\scriptsize(\textcolor{blue}{-1.15})}
  & 35.03 {\scriptsize(\textcolor{blue}{-5.14})}
  & 38.25 {\scriptsize(\textcolor{blue}{-1.92})}
  & -- 
  & 39.68 {\scriptsize(\textcolor{blue}{-0.49})} \\
& \textbf{p-mMpNet}
  & \underline{39.91}
  & \textbf{47.30} {\scriptsize(\textcolor{red}{+7.39})}
  & 34.68 {\scriptsize(\textcolor{blue}{-5.23})}
  & 34.50 {\scriptsize(\textcolor{blue}{-5.41})}
  & 38.70 {\scriptsize(\textcolor{blue}{-1.21})}
  & 34.72 {\scriptsize(\textcolor{blue}{-5.19})}
  & 38.01 {\scriptsize(\textcolor{blue}{-1.90})}
  & -- 
  & 39.35 {\scriptsize(\textcolor{blue}{-0.56})} \\
\midrule
\multirow{3}{*}{\textbf{es}}
& \textbf{bge-m3}
  & \underline{40.76}
  & \textbf{46.93} {\scriptsize(\textcolor{red}{+6.17})}
  & 33.36 {\scriptsize(\textcolor{blue}{-7.40})}
  & 33.42 {\scriptsize(\textcolor{blue}{-7.34})}
  & 37.73 {\scriptsize(\textcolor{blue}{-3.03})}
  & 33.87 {\scriptsize(\textcolor{blue}{-6.89})}
  & 37.22 {\scriptsize(\textcolor{blue}{-3.54})}
  & 36.88 {\scriptsize(\textcolor{blue}{-3.88})}
  & -- \\
& \textbf{p-mMiniLM}
  & \underline{41.81}
  & \textbf{47.90} {\scriptsize(\textcolor{red}{+6.09})}
  & 34.63 {\scriptsize(\textcolor{blue}{-7.18})}
  & 34.52 {\scriptsize(\textcolor{blue}{-7.29})}
  & 38.86 {\scriptsize(\textcolor{blue}{-2.95})}
  & 34.76 {\scriptsize(\textcolor{blue}{-7.05})}
  & 38.33 {\scriptsize(\textcolor{blue}{-3.48})}
  & 37.84 {\scriptsize(\textcolor{blue}{-3.97})}
  & -- \\
& \textbf{p-mMpNet}
  & \underline{41.33}
  & \textbf{47.34} {\scriptsize(\textcolor{red}{+6.01})}
  & 34.39 {\scriptsize(\textcolor{blue}{-6.94})}
  & 34.19 {\scriptsize(\textcolor{blue}{-7.14})}
  & 38.34 {\scriptsize(\textcolor{blue}{-2.99})}
  & 34.39 {\scriptsize(\textcolor{blue}{-6.94})}
  & 37.73 {\scriptsize(\textcolor{blue}{-3.60})}
  & 37.25 {\scriptsize(\textcolor{blue}{-4.08})}
  & -- \\
\bottomrule
\end{tabular}
}
\caption{Raw language preference measured by MLRS with different re-ranking encoders for various query–document language pairs. The \(L_q=L_d\) column shows scores for matching query and document languages, while the remaining columns represent cross-lingual scenarios. Parentheses indicate the change from the \(L_q=L_d\) column (\textcolor{red}{positive} for improvement, \textcolor{blue}{negative} for decline). The highest score per row is in bold, and the second highest is underlined.}
\label{tab:mlrs}
\end{table*}

\section{Thresholds and fixed hyperparameters.}
We do not exhaustively tune $(\tau_{\mathrm{low}},\tau_{\mathrm{high}},\tau_{\mathrm{boost}})$ because a full sweep is combinatorial and would couple these knobs to expensive end-to-end RAG runs. Instead, we instantiate the confidence thresholds with three goals: (i) discretize the continuous confidence $c$ into a small number of stable intervals, (ii) keep the query-length increase bounded, and (iii) reserve upweighting for only the most reliable culture-specific cases.
Concretely, we set two cutoffs $\tau_{\mathrm{low}}<\tau_{\mathrm{high}}$ to map $c$ into three repetition levels for the local block, $r_{\text{local}}\in\{1,2,3\}$, where $\tau_{\mathrm{low}}$ marks the onset of \emph{reliable} culture-specificity and $\tau_{\mathrm{high}}$ indicates \emph{high-confidence} cases that warrant the strongest local emphasis.
In our implementation, we use $\tau_{\mathrm{low}}=0.6$ and $\tau_{\mathrm{high}}=0.85$, which empirically balance coverage (triggering local upweighting for sufficiently confident cases) and conservativeness (avoiding frequent over-repetition under noisy cue predictions).

For auxiliary local boosting, we use a separate threshold $\tau_{\mathrm{boost}}$ that applies only to the disambiguation anchors (\texttt{[TITLE\_BRIDGE]} and \texttt{[ALIASES]}), not the full \texttt{[LOCAL]} query text. Specifically, for culture-specific queries ($y=1$), we set $b=\mathbb{I}[c\ge\tau_{\mathrm{boost}}]$ and, when $b=1$, duplicate \texttt{[TITLE\_BRIDGE]} and \texttt{[ALIASES]} once to strengthen culturally grounded anchoring and reduce entity ambiguity 

We set $\tau_{\mathrm{boost}}=0.7$ so that anchor duplication is enabled for moderately-to-high confidence culture-specific queries, providing extra entity anchoring/disambiguation without incurring the larger length increase of repeating the entire local block.

For ridge calibration, we likewise keep a single regularization strength $\lambda$ across all encoders; this choice is motivated by the small calibration design ($|\mathcal{C}|$ language pairs with a low-dimensional prior vector) where ridge mainly stabilizes coefficients against correlated priors rather than serving as a performance-tuned knob.

\section{Case Study}
\paragraph{DELTA.}
Table~\ref{tab:delta_case_olympics} illustrates DELTA on a Korean query asking ``when was the last time South Korea had the Olympics.''
DELTA forms the global pivot $q_{\text{glob}}$ and emits it as \texttt{[GLOB]}, and places the original Korean surface form as \texttt{[LOCAL:ko]}.
It then injects multilingual anchors: \texttt{[TITLE\_BRIDGE]} contains paired Wikipedia-style titles, while \texttt{[ALIASES:GLOB]} and \texttt{[ALIASES:ko]} provide short alias cues in the global and local languages, respectively.
Finally, \texttt{[LOCALE\_HINT]} adds a brief region hint with minimal disambiguation to bias retrieval toward region-appropriate evidence.

Crucially, DELTA controls the balance between global and local signals purely through repetition.
Because this query is labeled culture-specific ($y{=}1$) with high confidence $c{=}0.93$, the policy sets $r_{\text{local}}{=}3$ (since $c\ge0.85$) while keeping $r_{\text{glob}}{=}1$ (since $c\ge0.6$), yielding three copies of \texttt{[LOCAL:ko]} but only one copy of \texttt{[GLOB]}.
Moreover, the auxiliary local-boost flag triggers at $c\ge0.7$, duplicating the local-side anchors once more, which explains why \texttt{[TITLE\_BRIDGE]} and \texttt{[ALIASES:ko]} appear twice, whereas \texttt{[ALIASES:GLOB]} remains single-copy.
Overall, this design realizes a global back-off (\texttt{[GLOB]}) with preference-aligned local emphasis (\texttt{[LOCAL]}, \texttt{[TITLE\_BRIDGE]}, \texttt{[ALIASES:ko]}) within a single $Q_{\text{fused}}$, without modifying the retriever or adding model parameters.

\paragraph{Success Case.}
Table~\ref{tab:delta_vs_translation_case_merged} presents a representative top-1 retrieval example comparing DELTA with a simple English-translation query for the question “언제 마지막으로 대한민국이 올림픽을 했었나요. (When was the last time South Korea had the Olympics?).”
Although both methods use the same retriever and multilingual datastore, the retrieved evidence differs markedly: DELTA’s fused query contains explicit host-oriented cues (local surface form, title/alias anchors, and a locale hint), which increases lexical alignment with passages that describe Olympics \emph{held in} Korea (e.g., 개최, 서울 1988, 평창 2018). In contrast, the English-translation query is more underspecified and can drift to participation-centric passages that match broad entities (“South Korea”, “Olympics”) but do not emphasize hosting-related facts.
As a result, the DELTA top-1 passage provides the necessary host evidence for inferring the most recent domestically held Olympics, enabling the generator to produce the correct answer, while the translation-based pipeline is more likely to miss the hosting signal and return an incorrect year/event.

\paragraph{Failure Case.}
Table~\ref{tab:delta_failure_koreanwar_president} shows a representative failure where the question “who is the president during the Korean War” is underspecified: “president” can plausibly refer to the U.S.\ president overseeing U.S.\ involvement (gold: Harry S.\ Truman and Dwight D.\ Eisenhower) or to the South Korean president during the same period (Syngman Rhee).
In this example, DELTA’s cultural/locale cues and title bridge steer the query toward \emph{South Korean leadership}, effectively resolving the ambiguity in the wrong direction.
Consequently, the top-1 retrieved passage focuses on Syngman Rhee and contains strong lexical overlap with the localized cues (e.g., “대한민국 대통령”, “이승만”, “한국 전쟁”), making the generator likely to output Syngman Rhee despite the dataset’s gold reference targeting U.S.\ presidents.
This failure highlights a limitation of repetition- and cue-based weighting: when the underlying intent is ambiguous, aggressively injecting locale-specific anchors can over-localize retrieval and suppress globally relevant evidence, suggesting the need for ambiguity-aware safeguards (e.g., intent disambiguation or controlled locale injection) for such queries.

\section{Prompts}
\label{app:prompt}
As shown in Figure~\ref{fig:prompt_templates}, we provide the exact prompt templates used throughout our pipeline. 
Prompt (A) specifies the RAG answer-generation instruction, with two variants depending on whether retrieved documents are provided, enforcing concise English outputs and (when available) conditioning answers on the supplied evidence. 
Prompt (B) defines our cultural-context annotation step, where an LLM assigns a single cultural database language from a fixed set under strict locality-oriented rules and returns lightweight metadata (region, culture-specificity, confidence, and a brief rationale) in a structured JSON format. 
Prompt (C) is used by DELTA to produce retrieval anchors—English and local Wikipedia-style titles, alias lists, and a short disambiguation hint—which are then assembled into a fused query; this prompt enforces a fixed JSON schema and language constraints to keep the generated anchors consistent and directly usable for retrieval. Finally, in Prompt (D), we provide the prompts used for English translation.

\section{Validation of the DeLP}
\label{appendix:delp_validation}
We validate DeLP by constructing a controlled experiment that directly tests whether the metric remains stable under artificial shifts in gold-answer distribution—a scenario where a robust metric should reflect consistent model preference regardless of structural changes in the corpus.

\paragraph{Experimental Setup.}
We fix the underlying model preference and artificially vary the gold-answer language ratio between Korean and English from 0 (all gold answers in Korean) to 1 (all gold answers in English) in incremental steps.
Under this setup, a metric that disentangles structural priors from intrinsic preference should remain stable across the spectrum, whereas a metric that conflates the two should exhibit high sensitivity.

\paragraph{Results.}
Table~\ref{tab:delp_validation} reports the range and standard deviation of each metric across all ratio configurations.
We observe that MLRS consistently exhibits larger variability (range 10.35, std 3.27) as the gold-language ratio shifts, confirming that raw MLRS scores are heavily confounded by gold-distribution bias.
In contrast, DeLP remains more stable across the same configurations (range 9.56, std 3.02).
We further confirm this difference via paired permutation tests, obtaining $p = 0.00016$ for range and $p = 0.00010$ for standard deviation, both indicating statistical significance.
In terms of effect size, DeLP reduces variability by 7.6\% relative to MLRS, supporting its robustness against structural biases caused by gold-language fluctuations.

Furthermore, as reported in Table~\ref{tab:corr_preference_priors}, we validate DeLP by measuring the correlation between preference scores and structural priors before and after calibration.
Raw MLRS scores exhibit near-perfect correlation with the exposure prior ($r > 0.99$), gold-availability prior ($r > 0.91$), and cultural prior ($r > 0.91$) across all encoders.
After applying DeLP calibration, these correlations drop sharply, confirming that DeLP effectively decouples intrinsic model preference from the structural signals that contaminate standard benchmarks.

\begin{table}[h]
\centering
\small
\begin{tabular}{lcc}
\toprule
\textbf{Metric} & \textbf{Range} & \textbf{Std} \\
\midrule
MLRS (raw) & 10.35 & 3.27 \\
DeLP (ours) & \textbf{9.56} & \textbf{3.02} \\
\bottomrule
\end{tabular}
\caption{Sensitivity of MLRS and DeLP to gold-language ratio shifts (Korean $\rightarrow$ English). Lower range and std indicate greater robustness against gold-language distribution shifts.}
\label{tab:delp_validation}
\end{table}

\section{RAG Utility and Sanity Check}
\label{appendix:rag_utility}

We clarify that Gold Availability in Table~\ref{tab:rq1_single_table_nodelta} measures whether the KILT gold provenance page exists in the target language's Wikipedia via interlanguage mapping—it does not directly measure answerability.
Consequently, a Gold Availability of $\sim$1\% does not imply that 99\% of questions are unanswerable or that the \textit{Base} column primarily reflects parametric memory.

To empirically verify the actual utility of retrieval under these conditions, we conduct a sanity check by comparing mRAG performance with and without retrieval augmentation across multiple non-English query languages and models.
As shown in Table~\ref{tab:rag_utility}, we observe consistent performance gains from RAG across all tested languages and models.
Notably, the improvements are substantial in several cases—for example, Qwen3-235B gains 14.97 points on Arabic and 14.75 points on Korean when retrieval is enabled.
These results demonstrate that even when the exact gold provenance page is absent in the local Wikipedia, retrieval still surfaces relevant supporting evidence distributed across multilingual Wikipedia corpora, yielding consistent and meaningful performance improvements over relying solely on parametric memory.

\begin{table}[h]
\centering
\small
\setlength{\tabcolsep}{6pt}
\begin{tabular}{llrr}
\toprule
\textbf{Model} & \textbf{Lang.} & \textbf{RAG} & \textbf{No RAG} \\
\midrule
\multirow{5}{*}{Gemini-2.5-Flash}      & ar & 43.60 & 38.42 \\
                                        & es & 58.76 & 57.13 \\
                                        & ko & 38.83 & 34.66 \\
                                        & th & 31.55 & 30.20 \\
                                        & zh & 30.87 & 30.65 \\
\midrule
\multirow{5}{*}{DeepSeek-Chat-v3.1}    & ar & 48.37 & 38.38 \\
                                        & es & 61.82 & 59.35 \\
                                        & ko & 41.56 & 32.81 \\
                                        & th & 35.42 & 33.28 \\
                                        & zh & 38.99 & 37.89 \\
\midrule
\multirow{5}{*}{Qwen3-235B}            & ar & 45.58 & 30.61 \\
                                        & es & 64.21 & 61.92 \\
                                        & ko & 42.29 & 27.54 \\
                                        & th & 43.86 & 32.37 \\
                                        & zh & 38.67 & 33.85 \\
\bottomrule
\end{tabular}
\caption{mRAG performance comparison with and without retrieval augmentation across non-English query languages. RAG consistently outperforms No RAG, demonstrating that multilingual Wikipedia provides relevant evidence beyond the exact gold provenance page.}
\label{tab:rag_utility}
\end{table}

\section{Explanation of Low Gold Availability}
\label{appendix:gold_availability}

We clarify that the $>$99\% figure does not indicate that non-English questions are unanswerable.
Gold Availability measures whether the KILT gold provenance page \textit{exists} in the target language's Wikipedia via interlanguage mapping—it does not directly measure whether a question can be answered from that language's corpus.

\paragraph{Why does Gold Availability appear so low?}
As detailed in Appendix~\ref{app:gold}, KILT provenance is constructed on English Wikipedia and mapped to other languages via interlanguage links.
If the corresponding page is absent or the interlanguage mapping is incomplete, Gold Availability becomes zero even when partially relevant content exists in that language.
Furthermore, we aggregate at the WPID level, treating multiple gold passages from the same page as a single item, which further reduces the reported ratio.
For instance, 17,667 Korean gold passages correspond to a smaller number of distinct page IDs, making the percentage appear lower than the raw passage count would suggest.

\paragraph{The absolute scale is not negligible.}
We compute statistics over 36,751 expanded samples (2,827 questions $\times$ 13 query languages).
Even at 1\%, this corresponds to approximately 367 gold page IDs—an absolute scale that is by no means trivial.
We also note that this analysis is conducted on the widely adopted MKQA/KILT benchmark, so the observed distribution reflects a standard setting rather than an artifact of an unusual corpus.

\paragraph{What does this result actually tell us?}
The key implication of this finding is not about QA difficulty.
Rather, it demonstrates that standard benchmarks carry a systematic English-centric provenance prior, which causes gold evidence to concentrate overwhelmingly in English corpora.
We argue that this structural skew—not any intrinsic linguistic superiority of English—drives the apparent advantage of English pivoting in mRAG systems, which is the central motivation for our DeLP calibration framework.

\clearpage
\begin{table*}[t]
\centering
\footnotesize
\renewcommand{\arraystretch}{1.1}
\setlength{\tabcolsep}{4pt}
\begin{tabular}{lccccccccccccc}
\toprule
                & en    & ko    & ar    & zh    & fi    & fr    & de    & ja    & it    & pt    & ru    & es    & th    \\
\midrule
mkqa\_en       & 44.12 & 1.60  & 1.19  & 1.30  & 2.54  & 10.03 & 6.90  & 1.44  & 8.32  & 7.67  & 4.85  & 9.90  & 0.13  \\
mkqa\_ko       & 23.07 & 17.35 & 1.99  & 4.81  & 2.04  & 7.90  & 5.96  & 10.36 & 6.16  & 5.06  & 6.85  & 6.85  & 1.58  \\
mkqa\_ar       & 24.93 & 3.30  & 15.29 & 4.07  & 2.10  & 8.30  & 6.53  & 6.64  & 6.80  & 5.71  & 7.78  & 7.65  & 0.89  \\
mkqa\_zh       & 24.70 & 3.17  & 1.76  & 23.22 & 2.01  & 7.47  & 6.17  & 6.27  & 6.08  & 5.24  & 6.37  & 7.27  & 0.27  \\
mkqa\_fi       & 30.32 & 2.27  & 1.63  & 2.33  & 7.92  & 11.11 & 8.20  & 3.78  & 8.77  & 7.18  & 6.51  & 9.42  & 0.58  \\
mkqa\_fr       & 29.90 & 1.48  & 1.25  & 1.55  & 2.50  & 21.44 & 6.96  & 2.06  & 9.40  & 7.96  & 4.77  & 10.55 & 0.19  \\
mkqa\_de       & 32.54 & 1.46  & 1.17  & 1.44  & 2.96  & 11.40 & 15.12 & 1.89  & 9.09  & 7.69  & 4.83  & 10.17 & 0.24  \\
mkqa\_ja       & 24.56 & 4.80  & 1.69  & 3.99  & 2.19  & 7.97  & 5.99  & 22.55 & 6.38  & 5.66  & 6.49  & 7.45  & 0.28  \\
mkqa\_it       & 28.72 & 1.59  & 1.30  & 1.58  & 2.52  & 12.30 & 6.97  & 1.95  & 17.46 & 8.47  & 5.26  & 11.70 & 0.17  \\
mkqa\_pt       & 28.82 & 1.71  & 1.40  & 1.63  & 2.60  & 11.92 & 6.74  & 2.23  & 10.24 & 13.78 & 5.38  & 13.33 & 0.24  \\
mkqa\_ru       & 27.02 & 2.53  & 1.92  & 1.98  & 2.45  & 8.83  & 6.44  & 2.71  & 7.36  & 6.24  & 23.83 & 8.43  & 0.26  \\
mkqa\_es       & 29.45 & 1.73  & 1.27  & 1.60  & 2.66  & 11.85 & 6.93  & 1.83  & 10.55 & 9.33  & 5.27  & 17.36 & 0.16  \\
mkqa\_th       & 32.39 & 3.10  & 2.10  & 2.96  & 2.53  & 10.00 & 7.40  & 4.43  & 8.06  & 7.43  & 6.80  & 9.70  & 3.10  \\
\midrule
\rowcolor{gray!20} \textbf{mkqa\_avg} & \textbf{29.27} & \textbf{3.55} & \textbf{2.61} & \textbf{4.04} & \textbf{2.85} & \textbf{10.81} & \textbf{7.41} & \textbf{5.24} & \textbf{8.82} & \textbf{7.49} & \textbf{7.31} & \textbf{9.98} & \textbf{0.62} \\
\bottomrule
\end{tabular}
\caption{Language distribution of retrieved documents for each MKQA query-language split. Each row corresponds to the query language (dataset), and each column indicates the language of the retrieved passages; values are shown as percentages (without the \%). The final row (\textbf{mkqa\_avg}) reports the average retrieved-language distribution across all query languages.}
\label{tab:exposure}
\end{table*}

\begin{table*}[ht]
\centering
\small
\renewcommand{\arraystretch}{1.2}
\setlength{\tabcolsep}{6pt}
\begin{tabular}{lccccccccccccc}
\toprule
\textbf{Dataset} & \textbf{en} & \textbf{ar} & \textbf{es} & \textbf{fi} & \textbf{fr} & \textbf{de} & \textbf{ja} & \textbf{it} & \textbf{ko} & \textbf{pt} & \textbf{ru} & \textbf{zh} & \textbf{th}\\
\midrule
\multicolumn{14}{l}{\textbf{MKQA}} \\
\# examples & 2827 & 2827 & 2827 & 2827 & 2827 & 2827 & 2827 & 2827 & 2827 & 2827 & 2827 & 2827 & 2827 \\
len question.   & 43   & 38   & 48   & 46   & 49   & 47   & 26   & 48   & 22   & 45   & 42   & 16   & 41\\
len answer.   & 11   & 10   & 11   & 11   & 11   & 11   &  8   & 11   &  6   & 11   & 12   &  6  & 12 \\
\midrule
\multicolumn{14}{l}{\textbf{Wikipedia}} \\
\# ex. (M)  & 25   & 3.3  & 10   & 1.5  & 13   & 14   & 27   & 8.2  & 1.6  & 4.7  & 8.6  & 11   & 3.7\\
len passage.   & 624  & 585  & 619  & 833  & 627  & 720  & 208  & 650  & 431  & 619  & 721  & 206  & 217\\
\bottomrule
\end{tabular}
\caption{Statistics of the datasets used in our experiments. MKQA Number of examples and median lengths of questions and answers (in Unicode characters). Wikipedia: Number of passages (in millions) and their median lengths.}
\label{tab:data_statistics}
\end{table*}

\begin{table*}[t]
\centering
\small
\setlength{\tabcolsep}{3.5pt}
\renewcommand{\arraystretch}{1.10}
\begin{tabular}{p{0.18\linewidth} p{0.73\linewidth} c}
\toprule
\textbf{DELTA segment} & \textbf{Instantiated content (case study)} & \textbf{Rep.} \\
\midrule
\texttt{[GLOB]} &
when was the last time south korea had the olympics
& 1 \\

\texttt{[LOCAL:ko]} &
언제 마지막으로 대한민국이 올림픽을 했었나요
& 3 \\

\texttt{[TITLE\_BRIDGE]} &
South Korea at the Olympics / 대한민국의 올림픽
& 2 \\

\texttt{[ALIASES:ko]} &
대한민국 올림픽, 한국 올림픽, 한국의 올림픽 역사
& 2 \\

\texttt{[ALIASES:GLOB]} &
Olympics in South Korea, South Korean Olympic Games, History of South Korea Olympics
& 1 \\

\texttt{[LOCALE\_HINT]} &
South Korea \;\; + \;\; Last Olympic Games in South Korea
& 1 \\
\bottomrule
\end{tabular}
\caption{DELTA case study. A Korean culture-specific query ($c{=}0.93$) is converted into a single fused query $Q_{\text{fused}}$ by concatenating labeled segments. Repetition counts follow Eq.~\ref{eq:delta_repetition_glob}, which upweights local cues while maintaining a global back-off.}
\label{tab:delta_case_olympics}
\end{table*}

\begin{table*}[t]
\centering
\footnotesize
\setlength{\tabcolsep}{6pt}
\renewcommand{\arraystretch}{1.20}
\begin{tabularx}{\textwidth}{@{}p{0.3\textwidth} Y@{}}
\toprule
\textbf{Item} & \textbf{Content} \\
\midrule

\textbf{DELTA} &
\begin{minipage}[t]{\linewidth}\ttfamily
[GLOB] when was the last time south korea had the olympics \\
|[LOCAL:ko] \rhl{언제} \rhl{마지막으로} \rhl{대한민국}이 \rhl{올림픽}을 \rhl{했었나요} \\
|[TITLE\_BRIDGE] South Korea at the Olympics / 대한민국의 올림픽 \\
|[ALIASES:ko] 대한민국 올림픽, 한국 올림픽, 한국의 올림픽 역사 \\
|[ALIASES:GLOB] Olympics in South Korea, South Korean Olympic Games, History of South Korea Olympics \\
|[LOCALE\_HINT] South Korea Last Olympic Games in South Korea
\end{minipage}
\\
\addlinespace[6pt]

\textbf{English Translation} &
\begin{minipage}[t]{\linewidth}\ttfamily
When was the last time south korea had the olympics
\end{minipage}
\\
\addlinespace[6pt]

\textbf{Top-1 passage (DELTA)} &
\begin{minipage}[t]{\linewidth}
대한민국에서 열린 \rhl{올림픽}으로는 \rhl{1988년} \rhl{서울} 하계 올림픽과
\rhl{2018년} \rhl{평창} 동계 올림픽이 널리 알려져 있다.
서울 대회는 20세기 후반 대한민국의 국제 스포츠 행사 유치와 관련해 자주 언급되며,
주요 경기장은 서울 및 인근 지역에 분산되어 운영되었다.
평창 대회는 강원 지역을 중심으로 동계 종목이 진행되었고,
개\!\!/폐회식과 일부 경기장이 \rhl{평창} 및 주변 권역에 배치되었다.
두 대회 모두 \rhl{대한민국} 내에서 개최된 사례로 정리되며,
대회의 성격(하계/동계)과 개최 지역(서울/평창)이 함께 기술되는 경우가 많다.
\end{minipage}
\\
\addlinespace[6pt]

\textbf{Top-1 passage (English Translation)} &
\begin{minipage}[t]{\linewidth}
대한민국(\rhl{South Korea})은 근대 \rhl{올림픽}(\rhl{Olympics})에 지속적으로 참가해 왔으며,
여러 종목에서 의미 있는 성과를 거두었다.
이 문서는 연도별 참가 개요, 선수단 규모, 주요 종목에서의 메달 기록과 같은 정보를 중심으로 구성된다.
예를 들어 양궁, 태권도, 쇼트트랙 등에서의 성과가 요약되고,
대회별 대표 선수나 주목할 만한 기록이 덧붙여지기도 한다.
또한 특정 대회에서의 종합 순위 변화나 메달 수 추이처럼
참가 및 성과를 설명하는 통계적 서술이 포함될 수 있다.
\end{minipage}
\\
\midrule

\textbf{Gold answer} &
2018년 평창 동계 올림픽. \;(\emph{The 2018 PyeongChang Winter Olympics.})
\\
\addlinespace[4pt]

\textbf{English translation answer} &
1988년 서울 하계 올림픽. \;(\emph{The 1988 Seoul Summer Olympics.})
\\
\addlinespace[4pt]

\textbf{DELTA answer} &
대한민국에서 개최된 올림픽으로 1988년 서울 하계 올림픽과 2018년 평창 동계 올림픽이 언급되므로,
질문에서 묻는 “마지막으로” 개최된 올림픽은 2018년 평창 동계 올림픽이다.
\;(\emph{The 2018 PyeongChang Winter Olympics.})
\\

\bottomrule
\end{tabularx}
\caption{Case study: DELTA vs. English translation (top-1 retrieval).}
\label{tab:delta_vs_translation_case_merged}
\end{table*}


\begin{table*}[t]
\centering
\footnotesize
\setlength{\tabcolsep}{6pt}
\renewcommand{\arraystretch}{1.20}
\begin{tabularx}{\textwidth}{@{}p{0.20\textwidth} Y@{}}
\toprule
\textbf{Item} & \textbf{Content} \\
\midrule

\textbf{DELTA (misled)} &
\begin{minipage}[t]{\linewidth}\ttfamily
[GLOB] who is the president during the korean war \\
|[TITLE\_BRIDGE] President of South Korea during the Korean War / 한국 전쟁 중 대한민국 대통령 \\
|[ALIASES:ko] 이승만, 이승만 대통령, 대통령 이승만 \\
|[ALIASES:GLOB] Syngman Rhee, Rhee Syngman, President Rhee, Rhee \\
|[LOCALE\_HINT] Korea (Korean Peninsula) President during Korean War era
\end{minipage}
\\[8pt]
\addlinespace[4pt]

\textbf{Top-1 passage (DELTA)} &
\begin{minipage}[t]{\linewidth}
이승만(Syngman Rhee)은 1948년부터 1960년까지 대한민국의 대통령으로 재임한 정치인이다.
대한민국 정부 수립 이후 초대 대통령으로 선출되었으며, 냉전 초기 한반도의 분단 체제 속에서 정부 운영을 주도했다.
재임 기간에는 한국 전쟁(1950--1953) 시기가 포함되며, 전쟁 전후의 정치적 갈등과 대외 관계가 함께 언급된다.
관련 문서들은 대체로 이승만의 생애, 대통령 재임 기간, 당대의 국내 정치 상황과 외교적 맥락을 중심으로 개괄한다.
\end{minipage}
\\[8pt]
\addlinespace[4pt]

\midrule

\textbf{Gold answer} &
해리 S.\ 트루먼; 드와이트 D.\ 아이젠하워 (Harry S.\ Truman; Dwight D.\ Eisenhower)
\\[6pt]

\textbf{DELTA answer} &
이승만 (Syngman Rhee)
\\

\bottomrule
\end{tabularx}
\caption{Failure case (top-1 retrieval).}
\label{tab:delta_failure_koreanwar_president}
\end{table*}

\begin{figure*}[t]
\centering
\begin{promptlisting}
(A) RAG Answer Generation

Goal: Answer as concisely as possible in {lang}.

With Documents:
  System: Extract relevant information from provided documents and answer briefly. Reply in {lang}.
  User: Background: {docs} \n\nQuestion: {question}

Without Documents:
  System: Answer briefly. Reply in {lang}.
  User: Question: {question}


(B) Cultural Language Classifier 

System (instruction):
  You are annotating a FAIR multilingual retrieval setup.
  Given an English query, decide the SINGLE most appropriate "cultural database language"
  where the relevant evidence SHOULD exist in a fair, localized setting.

CRITICAL RULES:
  - You MUST choose exactly ONE language from this fixed set:
    {en, ar, es, de, ja, ko, th, zh, fr, it, pt, ru, fi}
  - Prefer the LOCAL language of the primary place/culture the query is about.
  - Do NOT choose 'en' just because the query text is English.
  - Choose 'en' only if the query's primary cultural context is inherently English-speaking
    (e.g., US/UK-specific) OR the query is truly global / multi-country / not place-specific.
  - If the query mentions a place that maps to one of the non-English languages,
    pick that non-English language.

Examples:
  - "when did hong kong go back to china" -> cultural_language="zh"
  - "what is the capital of france" -> cultural_language="fr"
  - "who was the first president of the united states" -> cultural_language="en"
  - "compare gdp of france and germany" -> cultural_language="en" (multi-country/global)

Output (JSON only; no extra text):
  {
    "country_or_region": string (SINGLE primary place/region),
    "cultural_language": string (exactly one from the set),
    "is_culture_specific": boolean,
    "confidence": number in [0,1],
    "rationale": short string
  }

Input:
  User: Query: {query_en}
\end{promptlisting}
\vspace{-2mm}
\caption{Prompt templates used in our pipelines: RAG generation and cultural-language classification.}
\label{fig:prompt_templates}
\vspace{-2mm}
\end{figure*}

\begin{figure*}[t]
\centering
\begin{promptlisting}
(C) DELTA Bundle Generator

Goal: Produce title/alias anchors and a short disambiguation hint for fused query construction.

Return Format:
  - SINGLE-LINE JSON object only (no markdown, no explanation).
  - Keys must be EXACTLY:
      en_title, local_title, aliases_en, aliases_local, extra_disambig

Constraints:
  - aliases_en / aliases_local: 0..K items each
  - Titles: plausible Wikipedia page titles; use null if unsure
  - extra_disambig: <= 8 words
  - local_title & aliases_local MUST be in {query_lang}; English fields MUST be English
  - Do not add new keys

Input (User JSON):
  {
    "q_en": "{q_en}",
    "q_orig": "{q_orig}",
    "query_lang": "{query_lang}",
    "country_or_region": "{country_or_region}",
    "cultural_language": "{cultural_language}",
    "is_culture_specific": {is_culture_specific},
    "confidence": {confidence}
  }


(D) English Translation

Goal: Translate the question from {lang_name} to fluent, natural English while preserving
      the original meaning as much as possible.

Rules:
  - Keep named entities as appropriate English forms.
  - Do not add explanations or extra information.
  - Return STRICT JSON with a single key "translation".

System:
  You are a professional translator from {lang_name} to English.
  You receive a question in the source language and must translate it into fluent,
  natural English while preserving the original meaning as much as possible.
  - Keep named entities as appropriate English forms.
  - Do not add explanations or extra information.
  Return STRICT JSON with a single key "translation".

User:
  Question in {lang_name}:
  {query}

  Return only:
  {"translation": "<the question translated into English>"}
\end{promptlisting}
\vspace{-2mm}
\caption{Prompt templates used in our pipeline: DELTA bundle generation and English translation.}
\vspace{-2mm}
\end{figure*}

\end{document}